\documentclass[12pt]{article}
\usepackage{amsmath}
\usepackage{graphicx,psfrag,epsf}
\usepackage{enumerate}
\usepackage{natbib}
\usepackage{url} 
\usepackage{amsfonts}
\usepackage{verbatim}
\usepackage{ntheorem}
\usepackage{multirow}
\usepackage{extarrows}
\usepackage{algorithm}

\usepackage[noend]{algpseudocode}

\makeatletter
\def\algbackskip{\hskip-\ALG@thistlm}
\makeatother

\newtheorem{theorem}{Theorem}

\newcommand{\blind}{1}

\begin{document}

\def\spacingset#1{\renewcommand{\baselinestretch}%
{#1}\small\normalsize} \spacingset{1}


\if1\blind
{
  \title{\bf An Instrumental Variable Approach to Confounded Off-Policy Evaluation} 
\author{
Yang Xu$^1$, Jin Zhu$^2$, Chengchun Shi$^3$, Shikai Luo$^4$ and Rui Song$^1$\\
\textit{$^1$North-Carolina State University}\\
\textit{$^2$Sun Yat-sen University}\\
\textit{$^3$London School of Economics and Political Science}\\
\textit{$^4$ByteDance}\\
}
\date{\empty}
  \maketitle
} \fi

\if0\blind
{
  \bigskip
  \bigskip
  \bigskip
  \begin{center}
    {\LARGE\bf Doubly Robust Confounded Reinforcement Learning by Instrumental Variables}
\end{center}
  \medskip
} \fi

\bigskip

\begin{abstract}
Off-policy evaluation (OPE) is a method for estimating the return of a target policy using some pre-collected observational data generated by a potentially different behavior policy. In some cases, there may be unmeasured variables that can confound the action-reward or action-next-state relationships, rendering many existing OPE approaches ineffective. This paper develops an instrumental variable (IV)-based method for consistent OPE in confounded Markov decision processes (MDPs). Similar to single-stage decision making, we show that IV enables us to correctly identify the target policy's value in infinite horizon settings as well. Furthermore, we propose an efficient and robust value estimator and illustrate its effectiveness through extensive simulations and analysis of real data from a world-leading short-video platform. 
\end{abstract}

\noindent%
{\it Keywords: Instrumental Variables, Off-Policy Evaluation, Infinite-Horizons, Unmeasured Confounding, Reinforcement Learning.}  
\vfill

\newpage
\spacingset{1.45} 

\section{Introduction}\label{sec:intro}
Offline policy evaluation (OPE) estimates the discounted cumulative reward following a given target policy 
with an offline dataset collected from another (possibly unknown) behavior policy. 
OPE is important in situations where it is impractical or too costly to directly evaluate the target policy via online experimentation, including robotics \citep{quillen2018deep}, 
precision medicine \citep{murphy2003optimal,kosorok2019precision, tsiatis2019dynamic}, economics, quantitative social science \citep{abadie2018econometric}, recommendation systems \citep{li2010contextual,kiyohara2022doubly}, etc.

Despite a large body of literature on OPE (see Section \ref{sec:related_work} for detailed discussions), many of them rely on the assumption of no unmeasured confounders (NUC), excluding the existence of unobserved variables that could potentially confound either the action-reward or action-next-state pair. This assumption, however, can be violated in some real-world applications such as healthcare and technological industries. 

Our paper is partly motivated by the need to evaluate the long-term treatment effects of certain app download ads from a short-video platform. At each time, the platform may bid with many other companies to show their own ads to potential consumers. Unmeasured confounding poses a significant challenge in this data generating process. This is because other companies may win the auction and it remains unknown which ad is ultimately shown to the consumer. In addition, if the competitor's ad is displayed, the consumer may download their app instead. This lack of observability violates the no unmeasured confounders assumption, making it difficult to evaluate the effects of the ads consistently.
Recently, IV-based methods have stood out as a powerful approach to to account for unmeasured confounding and measurement errors 
and have been applied in a range of studies \citep{angrist1996identification,aronow2013beyond, tchetgen2013alternative, ogburn2015doubly, wang2018bounded, qiu2021optimal}. However, these methods are typically used 
in a single-stage setting and cannot be directly applied to general sequential decision making which is commonly 
encountered in the RL literature.

To fill in this gap, we propose an IV-based approach to OPE in confounded sequential decision making. The advances and contributions of our proposal 
are multi-fold. 

\textbf{Firstly}, to the best of our knowledge, this is one of the first paper to systematically examine the use of IVs for policy evaluation in infinite or long-horizon settings. Our proposal covers a range of models, including  Markov decision processes with unmeasured confounders (MDPUCs), high-order MDPs with unmeasured confounders and POMDPs, allowing the Markov assumption to be potentially violated in different levels. Existing IV-based RL approaches are mainly designed for the purpose of policy optimization, not policy evaluation. Moreover, related studies either rely on the Markov assumption \citep{liao2021instrumental,li2021causal,fu2022offline} or finite horizon settings \citep{chen2021estimating} with a few decision stages. This narrows the scope of their findings. 

\textbf{Secondly}, 
when specialized to MDPUCs, we develop a doubly robust policy value estimator. This new estimator, as guaranteed by semiparametric theory \citep{tsiatis2006semiparametric}, achieves the efficiency bound and thus provides the most robust and efficient value estimate for OPE in confounded MDPs. 
Existing semiparametrically efficient estimators designed for MDPs \citep{kallus2022efficiently} are biased in our setting, due to the existence of unmeasured confounders. \textbf{Finally}, as illustrated in Section \ref{sec:realdata}, our proposal offers valuable insights in helping tech industries to make sequential decisions in online digital advertising  
to improve consumers' conversion rates.

The rest of this paper is summarized as follows. In Section \ref{sec:related_work}, we review other related papers in the literature. Section \ref{sec:prelim} introduces necessary notations and the underlying causal diagram, serving as a preliminary foundation for the rest of the paper. Section \ref{sec:identification} discusses the identifiability of the value function. In Section \ref{sec:estimation}, we present three types of estimators, the efficient influence function, as well as the detailed estimation process along with the corresponding theoretical guarantees. In Section \ref{sec:extension}, we further extend our work to high-order MDPs and POMDPs. We conduct simulation studies in Section \ref{sec:simulation} and provide a real data analysis in Section \ref{sec:realdata}. The proofs for our 
main Theorems
can be found in the Supplementary Material.

\section{Related Works}\label{sec:related_work}
\subsection{Off-policy Evaluation}\label{sec:relatedOPE}
Over the past decades, OPE has been thoroughly researched in reinforcement learning  \citep[see][for an overview]{uehara2022review}. 
Current estimators can be roughly divided into three categories. The first type is the direct method estimator (DM) which 
directly constructs the policy value estimator via an estimated Q- or value function  
\citep{lagoudakis2003least, le2019batch, feng2020accountable, luckett2020estimating, hao2021bootstrapping, liao2021off, Chen2022on}. The second type is the importance sampling (IS)-based estimator that uses the (marginal) IS ratio to account for the distributional shift between the target and behavior policies \citep{thomas2015high, hallak2017consistent, hanna2017bootstrapping, liu2018breaking, schlegel2019importance, xie2019towards, dai2020coindice,   zhang2020gendice}. 
The last type 
combines DM and IS for robust OPE \citep{jiang2016doubly, thomas2016data, farajtabar2018more, Tang2020Doubly, uehara2020minimax, shi2021deeply, liao2022batch, kallus2022efficiently}. However, none of the aforementioned methods can handle unmeasured confounding. 

\subsection{Unmeasured Confounding}
In observational studies, the no unmeasured confounders (NUC) assumption is often violated due to the presence of latent variables. Recently, there has been an increasing focus on developing RL methods in confounded contextual bandits and sequential decision to address this problem. Some related references in confounded contextual bandits include \citet{bareinboim2015bandits,sen2017contextual,miao2018identifying,cui2020semiparametric,shi2020multiply,kallus2021causal,xu2021deep}. 
In general sequential settings, existing works can be broadly grouped into three categories. The first category of work relies on the Markov assumption, models the observed data via a confounded MDP \citep[MDPUC,][]{zhang2016markov}, and utilizes optimal balancing or certain proxy variables to handle the memoryless unobserved confounding \citep{bennett2021off,liao2021instrumental,wang2021provably,shi2022off,fu2022offline}. The second category uses a confounded partially observable MDP (POMDP) for problem formulation, borrows the idea from proximal causal inference \citep[see e.g.,][for an overview]{tchetgen2020introduction} and extends the framework to sequential decision making \citep{tennenholtz2020off,bennett2021proximal,nair2021spectral,miao2022off,shi2022minimax}. The last category develops partial identification bounds
for policy learning and evaluation based on sensitivity analysis \citep{kallus2020confounding,namkoong2020off,chen2021estimating}. 

\subsection{POMDPs}\label{sec:relatedPOMDP}
Our work is also closely related to a line of works on policy learning and evaluation in unconfounded POMDPs \citep{boots2011closing,anandkumar2014tensor,guo2016pac, azizzadenesheli2016reinforcement,jin2020sample,2021arXiv211012343H,kwon2021rl}. However, all the aforementioned methods are developed under settings without unmeasured confounders and are not directly applicable to our problem. Meanwhile, methods designed for confounded POMDPs require the action to be independent of the observation given the latent state \citep[see e.g.,][]{tennenholtz2020off,shi2022minimax}, which are not applicable to settings when the behavior policy depends on both the state and the observation.

\section{Preliminaries}\label{sec:prelim}
To illustrate the idea, we start by working with the MDPUC setup where the Markov assumption is satisfied. Extensions to non-Markov settings will be discussed in Section \ref{sec:extension}. 

Consider a single data trajectory where
$(S_t,A_t,R_t)$ 
denotes the state-action-reward triplet observed at time $t$. 
In the context of online digital advertising, both the action and the reward are binary variables. 
We denote $A_t=1$ if the ad is indeed exposed to the consumer at time $t$, and $R_t=1$ if the consumer is converted, i.e., downloaded our app at time $t$. Let $U_t$ denote the unobserved confounders at time $t$ which may affect both the action and reward/next state. In this example, $U_t$ includes the bidding strategies of other companies, as well as the information about the ad that is displayed to the consumer when $A_t=0$. $S_t$ is a vector which contains both the consumer's baseline information and the behavioral data (e.g., the number of historical requests of consumers from different media channels).

As we have mentioned in the introduction, the bidding strategies of other companies can impact both the ad exposure $A_t$ and the consumer's conversion rate $R_t$, resulting in a confounded dataset.
To address this problem, we leverage the IV (denoted by $Z_t$) to infer the long-term treatment effect. 
In our application, $Z_t$ is binary as well, depending on whether our company chooses to bid at time $t$ or not. We will illustrate in Section \ref{sec:realdata} that this is indeed a valid IV. 

\begin{figure}[t]
\centering
\includegraphics[width=16cm]{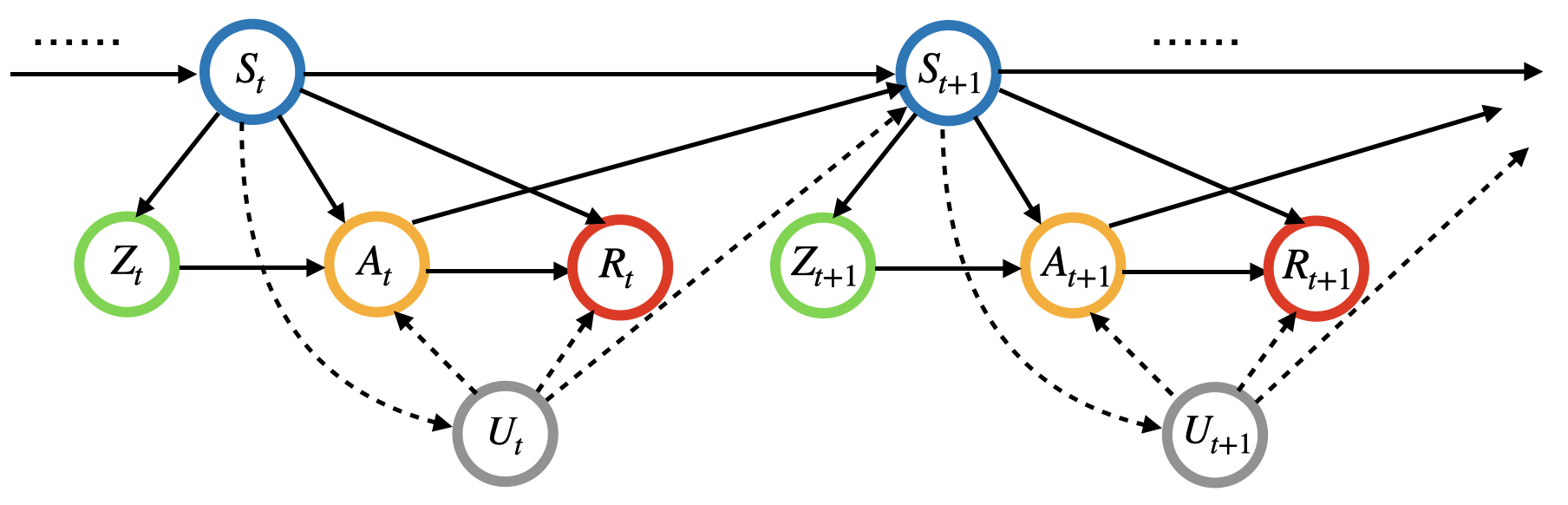}
\vspace{-10pt}
\caption{Causal diagram for IV-based MDPUC, where $U_t$ denotes the unmeasured confounders in between $A_t\rightarrow (R_t,S_{t+1})$.}\label{fig:causal_graph}
\end{figure}

To summarize, the complete data under the IV-based MDPUC model 
is given by $
\{(S_t,Z_t,A_t,R_t,U_t)\}_{t=0}^T$, where $T$ can be very large or infinite. A causal diagram depiciting the resulting data generating process is given in Figure \ref{fig:causal_graph}. 
The observed data contains $n$ i.i.d. trajectories, given by
\begin{equation}\label{eq:data}
   D_i=\{(S_{i,t},Z_{i,t},A_{i,t},R_{i,t})\}_{t=1}^T, \quad i=\{1,\dots,n\}. 
\end{equation}
Let $\pi: \mathcal{S}\times  \mathcal{A}\mapsto [0,1]$ denote the target policy we wish to evaluate, i.e., $\pi(a|s)=\mathbb{P}^{\pi}(A_t=a|S_t=s)$ for any $(a,s)\in \mathcal{S}\times  \mathcal{A}$. Likewise, let $b:\mathcal{S}\times \mathcal{U} \times \mathcal{A}\mapsto [0,1]$ denote the behaviour policy that generates the data in \eqref{eq:data}. Due to unmeasured confounding, the behavior policy is allowed to depend on both the observed state $S$ and the unobserved confounders $U$, and thus differs from $\pi$. 

For a given discounted factor $0\leq \gamma<1$, we define the value function $V^{\pi}(s_0)$ as the expected discounted sum of rewards starting from some initial state $s_0$ under policy $\pi$:
\begin{equation*}
    V^{\pi}(s_0)=\sum_{t=0}^{+\infty}\gamma^t\mathbb{E}^{\pi}(R_t|S_0=s_0),
\end{equation*}
where the superscript $\pi$ in $\mathbb{E}^{\pi}$ denotes the expectation of potential outcome of $R_t$ under policy $\pi$. 
We next define the aggregated value over the initial state distribution $\nu(s_0)$ as 
\begin{equation*}
\eta^{\pi}:=\mathbb{E}_{S_0\sim\nu}\big[V^{\pi}(S_0)\big].
\end{equation*}
Our objective lies in inferring $\eta^{\pi}$ based on \eqref{eq:data}. 

Directly applying existing OPE methods in Section \ref{sec:relatedOPE} will produce biased policy value estimators in the presence of unmeasured confounders. This is because  $\mathbb{E}^{\pi}(R_t|S_0)$ is generally not equal to $\mathbb{E}(R_t|S_0, A_j\sim \pi, 0\le j\le t)$. 
The former corresponds to the potential outcome generated by the causal diagram in Figure \ref{fig:causal_graph} with the arrows $\{U_t\rightarrow A_t\}_{0\leq t\leq T}$ removed, whereas the latter corresponds to the observed outcome 
generated under the original causal diagram in Figure \ref{fig:causal_graph}. 
This makes the identification and inference of $\eta^{\pi}$ become very tough to deal with. 

Before we conclude this section, let's summarize our model setup and the problem of interest. Using the data in (\ref{eq:data}), our goal is to efficiently estimate the outcome of executing a target policy $\pi$. In the subsequent sections, we will thoroughly examine the identification, estimation, and inference procedures for the value function $V^{\pi}(s_0)$ and aggregated value $\eta^{\pi}$ under confounded MDPs, high-order MDPs, as well as POMDPs.

\section{Identification}\label{sec:identification}
In this section, we show that the policy value can be consistently identified by Theorem \ref{thm:identification} below. Before we proceed, let's introduce the assumptions needed in the identification procedure.

We adopt a counterfactual outcome framework that is commonly used in the IV literature. Let $\bar{A}_t=(A_1,\dots,A_t)$ denote the action history up to time $t$, and $\bar{Z}_t=(Z_1,\dots,Z_t)$ denote the history of IVs up to time $t$. Define $A_t(\bar{z}_{t},\bar{a}_{t-1})$ as the potential action assigned to a subject at time $t$ if they were exposed to $\bar{Z}_t=\{\bar{z}_{t}\}$ and $\bar{A}_{t-1}=\{\bar{a}_{t-1}\}$, and $R_t(\bar{z}_{t},\bar{a}_{t})$, $S_{t+1}(\bar{z}_{t},\bar{a}_{t})$ as the potential reward and next state that would be observed if the subject were to receive $\{\bar{z}_{t}\}$ and $\{\bar{a}_t\}$ in the past.



\noindent\textbf{Assumption 1. (IV Assumptions)}\label{assump:1}\\
For any time $t\in\{1,\dots,T\}$, we assume: \\
\noindent(a) IV Independence: $Z_t\!\perp\!\!\!\perp U_t|S_t$.\\
\noindent (b) IV Relevance: $Z_t\not\!\perp\!\!\!\perp A_t|S_t
$.\\
\noindent (c) Exclusion Restriction: For any $\bar{z}_{t},\bar{a}_{t}$, $R_t(\bar{z}_{t},\bar{a}_{t})=R_t(\bar{z}_{t-1},\bar{a}_{t})$.\\
\noindent (d) $R_t(\bar{a}_{t}) \!\perp\!\!\!\perp(A_t,Z_t)|(S_t,U_t)$.\\
\noindent (e) Exclusion Restriction: For any $\bar{z}_{t},\bar{a}_{t}$, $S_{t+1}(\bar{z}_{t},\bar{a}_{t})=S_{t+1}(\bar{z}_{t-1},\bar{a}_{t})$. \\
\noindent (f) $S_{t+1}(\bar{z}_{t},\bar{a}_{t}) \!\perp\!\!\!\perp(A_t,Z_t)|(S_t,U_t)$.\\
\noindent (g) There is no additive $U-A$ interaction in both $\mathbb{E}[R_t(\bar{z}_{t},\bar{a}_{t})|S_t,U_t]$ and $\mathbb{E}[S_{t+1}(\bar{z}_{t},\bar{a}_{t})|S_t,U_t]$. That is,
\[
\begin{aligned}
&\mathbb{E}[R_t(\bar{z}_{t},\bar{a}_{t-1},a_t=1)-R_t(\bar{z}_{t},\bar{a}_{t-1},a_t=0)|S_t,U_t]\\
&=\mathbb{E}[R_t(\bar{z}_{t},\bar{a}_{t-1},a_t=1)-R_t(\bar{z}_{t},\bar{a}_{t-1},a_t=0)|S_t],\quad\\
\text{ and }\quad &\mathbb{E}[S_{t+1}(\bar{z}_{t},\bar{a}_{t-1},a_t=1)-S_{t+1}(\bar{z}_{t},\bar{a}_{t-1},a_t=0)|S_t,U_t]\\
&=\mathbb{E}[S_{t+1}(\bar{z}_{t},\bar{a}_{t-1},a_t=1)-S_{t+1}(\bar{z}_{t},\bar{a}_{t-1},a_t=0)|S_t].\quad
\end{aligned}
\]

Assumption 1 (a)-(c) ensure the validity of IVs, which are commonly used in the single-stage model setup \citep{angrist1995identification, abadie2003semiparametric, wang2018bounded, qiu2021optimal}. Assumption 1 (d), as discussed in \cite{wang2018bounded}, allows for common causes of $Z_t$ and $A_t$, and can be interpreted through d-separation. This assumption is mild in real-world settings, as it allows for common causes of $Z_t$ and $A_t$, $A_t$ and $(R_t,S_{t+1})$. Assumption 1 (e)-(f) is akin to (c)-(d), which ensures the impact of the IV to be the same for both the current-stage reward and next-stage state variables. As shown in the causal graph in Figure \ref{fig:causal_graph}, $R_t$ and $S_{t+1}$ have the same causal hierarchy, leading to similar IV-related assumptions. Assumption 1 (g) guarantees that conditioning on covariates $S_t$, unmeasured confounders $U_t$ only affect the causal effect of $A_t$ on the mean of current-state reward or next-state covariates in an additive way. This assumption is commonly used in related papers to ensure the indentifiability of the final estimand \citep{wang2018bounded, qiu2021optimal}.

Next, let's further impose the conditional independence assumptions that is commonly assumed in Markov decision processes. 
Define $\bar{W}_t$ as the set of all historical data up to stage $t$, where
\[
\bar{W}_t(\bar{z}_{t},\bar{a}_t)=\{S_0,U_0,R_0(z_0,a_0),\dots,S_{t}(\bar{z}_{t-1},\bar{a}_{t-1}),U_t,R_t(\bar{z}_{t},\bar{a}_{t})\}.
\] 
\noindent\textbf{Assumption 2. (Conditional Independence Assumptions)}\label{assump:2}\\
\noindent(a) {(MA)} Markov assumption: There exists a Markov transition
kernel $\mathcal{P}$ such that for any $t\geq 0$, $\bar{z}_{t}\in[0,1]^{t+1}$ and $\bar{a}_t\in[0,1]^{t+1}$, we have 
\[
\mathbb{P}(S_{t+1}(\bar{z}_{t},\bar{a}_{t})\in \mathcal{S}|\bar{W}_t(\bar{z}_{t},\bar{a}_t))=\mathcal{P}(\mathcal{S};z_t, a_t, S_t(\bar{z}_{t-1},\bar{a}_{t-1}),U_t).
\]
\noindent(b) {(CMIA)}  Conditional mean independence assumption: there exists a function $r$ such that for any $t\geq 0$, and $\bar{a}_t\in[0,1]^{t+1}$, we have 
\[
\mathbb{E}(R_{t}(\bar{z}_{t},\bar{a}_{t})|S_t(\bar{z}_{t-1},\bar{a}_{t-1}),\bar{W}_{t-1}(\bar{z}_{t-1},\bar{a}_{t-1}))=r( z_t,a_t, S_t(\bar{z}_{t-1},\bar{a}_{t-1}),U_t).
\]
\noindent(c) For any $t\in\{0,\dots, T\}$, the conditional distribution of $Z_t$, $A_t$ and $U_t$, given all historical data is only a function of the current state information. Specifically,
\[
\begin{aligned}
& \mathbb{E}(Z_{t}|S_t(\bar{z}_{t-1},\bar{W}_{t-1}(\bar{z}_{t-1},\bar{a}_{t-1}))=\mathbb{E}(Z_{t}|S_t(\bar{z}_{t-1})), \\
& \mathbb{P}(U_{t}|S_t(\bar{z}_{t-1},\bar{W}_{t-1}(\bar{z}_{t-1},\bar{a}_{t-1}))=\mathbb{P}(U_{t}|S_t(\bar{z}_{t-1})),\\
&\mathbb{E}(A_{t}(\bar{z}_{t},\bar{a}_{t-1})|S_t(\bar{z}_{t-1},\bar{a}_{t-1}),z_t,U_t,\bar{W}_{t-1}(\bar{z}_{t-1},\bar{a}_{t-1}))=\mathbb{E}(A_{t}(\bar{z}_{t},\bar{a}_{t-1})|S_t(\bar{z}_{t-1},\bar{a}_{t-1}),z_t,U_t).
\end{aligned}
\]

Assumption 2 is composed of a set of conditional independence assumptions, which require $\{Z_t,U_t,A_t,R_t,S_{t+1}\}$ to be independent of the past data history given the current-stage information. Similar assumptions are imposed in RL when NUC is satisfied \citep{ertefaie2014constructing,sutton2018reinforcement,luckett2020estimating}. 

It is worth mentioning that under Assumption 1 (c) and (e), we can further omit term $z_t$ on the RHS of all equations in Assumption 2. Moreover, when both Assumption 1 and 2 holds, the definition of $\bar{W}_t(\bar{z}_{t},\bar{a}_t)$ and the potential outcomes for $R_t$ and $S_{t+1}$ are a function of only $\bar{a}_t$, not $\bar{z}_t$. This result is easy to understand: Assumption 1 (c) restricts the effect of $z_t$ on $R_t$, making the potential outcome of $R_t$ independent of $z_t$ given the current-state action. Meanwhile, the conditional independence assumption ensures that $R_t$ won't be affected by the historical IVs $\bar{z}_{t-1}$, yielding the potential outcome of $R_t$ to be entirely independent of $\bar{z}_t$ given the action sequence $\bar{a}_t$. As such, one can relax some conditions in Assumption 1 without any loss of information. Details are provided in Proposition 1.
\vspace{0.1in}

\noindent\textbf{Proposition 1}.
\textit{Under Assumption 2, the exclusion restriction condition in Assumption 1 (c) is equivalent to assuming that $R_t(\bar{z}_{t},\bar{a}_{t})=R_t(\bar{a}_{t})$ holds for any $\bar{z}_t,\bar{a}_t$. Meanwhile, Assumption 1 (e) is equivalent to assuming that $S_{t+1}(\bar{z}_{t},\bar{a}_{t})=S_{t+1}(\bar{a}_{t})$ holds for any $\bar{z}_t,\bar{a}_t$.}

As we've discussed above, the proof of Proposition 1 is straightforward. Under Assumption 2 (b), 
\[
R_t(\bar{Z}_{t},\bar{A}_{t})\!\perp\!\!\!\perp \bar{Z}_{t-1} |(S_t,Z_t,A_t),
\]
which means that $R_t(\bar{z}_{t},\bar{a}_{t})=R_t({z}_{t},\bar{a}_{t})=R_t(\bar{a}_{t})$. The first equality holds by CIMA in Assumption 2 (b), and the second equality holds by the original exclusion restriction in Assumption 1 (c). Similarly, we can prove Assumption 1 (e) by only assuming that $S_{t+1}(\bar{z}_{t},\bar{a}_{t})=S_{t+1}(\bar{a}_{t})$ holds for any $\bar{z}_{t},\bar{a}_{t}$.

Finally, let's introduce the identification result based on the assumptions we imposed above.

\begin{theorem}\label{thm:identification} \textbf{(Identifiability)}\\
Under Assumptions 1-2, $V^{\pi}(s_0)$ equals 
\begin{equation}\label{eq:identification_V}
\begin{aligned}
\sum_{t,\tau_t}\gamma^tr_t
\bigg\{ \prod_{j=0}^t p_{r,s}(r_j,s_{j+1}|a_j,z_j,s_j) p_a(a_j|z_j,s_j)c(z_j|s_j)\bigg\},
\end{aligned}
\end{equation}
where $\tau_t:=\{z_j,a_j,r_j,s_{j+1}\}_{j=0}^t$ denotes the collection of all past $(z,a,r,s')$ tuples up to time $t$, and
\begin{equation}\label{eqn:ratio}
c(z_t|S_t)=\left\{\begin{aligned}
 \frac{p^{A}_1(S_t)-\pi(1|S_t)}{p^{A}_1(S_t)-p^{A}_0(S_t)}, &\quad\text{when } z_t=0 \\
\frac{\pi(1|S_t)-p^{A}_0(S_t)}{p^{A}_1(S_t)-p^{A}_0(S_t)},  &\quad\text{when } z_t=1
\end{aligned}\right.,
\end{equation}
in which $p^{A}_1(S_t):=\mathbb{E}[A_t|Z_t=1,S_t]$ and $p^{A}_0(S_t):=\mathbb{E}[A_t|Z_t=0,S_t]$. 
\end{theorem}

\noindent\textbf{Remark 1.} 
All the functions involved in \eqref{eq:identification_V} can be consistently estimated from the observed data, which thus implies the identifiability of $V^{\pi}(s_0)$. By taking expectation with respect to the initial state distribution, $\eta^{\pi}$ is also identifiable. 
Specifically,
\begin{equation*}
\small
\begin{aligned}   &\eta^{\pi}=\sum_{s_0}\nu({s_0})\cdot\Bigg[\sum_{t=0}^{T}\sum_{\{z_j,a_j,r_j,s_{j+1}\}_{j=0}^t}\gamma^tr_t\cdot\bigg\{ \prod_{j=0}^t p_{r,s}(r_j,s_{j+1}|a_j,z_j,s_j)\cdot p_a(a_j|z_j,s_j)\cdot c(z_j|s_j)\bigg\}\Bigg].   
\end{aligned}
\end{equation*}
\noindent\textbf{Remark 2.} The ratio function $c(z|s)$ in \eqref{eqn:ratio} measures the discrepancy between the behavior policy and the target $\pi$. In the special case where the target policy $\pi$ equals the behavior policy $b$, $c(z_t|S_t)$ is reduced to $p_z(z_t|S_t)$, i.e. the conditional probability density/mass function of $Z_t$ given $S_t$. In this case, it is immediate to see this equation holds since the product in the curly brackets of \eqref{eq:identification_V} corresponds to the joint probability density/mass function of the data trajectory up to time $t$. 
When $\pi\neq b$, $c(z|s)$ plays a similar role as the important sampling ratio to account for distributional shift. 

\noindent \textbf{Remark 3.} The main idea of the proof lies in first applying the conditional independence assumptions (Assumption 2) to decompose the cross-stage identification problem (i.e., $\mathbb{E}^{\pi} (R_t|S_0)$ for $t\ge 1$) into a sequence of single-stage problems, and then employ the IV-related conditions (Assumption 1) to replace the potential outcome distribution with the observed data distribution. More details about the proof can be found in Section \ref{appendix:ident} of the supplementary material.

\section{Estimation}\label{sec:estimation}
In this section, we discuss how to efficiently estimate $\eta^{\pi}$ under IV-based MDPUCs. We begin with introducing a direct method estimator and a marginal importance sampling estimator. Lastly, we present a doubly robust estimator,  which can be proved to be the most efficient in the presence of model misspecifications.

\subsection{Direct Method Estimator}\label{sec:DM}
We first introduce the DM estimator which constructs the policy value estimator based on an estimated Q-function. Toward that end, 
we define the Q-function in IV-based MDPUCs as
\begin{equation*}
    Q^{\pi}(s,z,a)=\mathbb{E}^{\pi}\bigg[\sum_{k=0}^{\infty}\gamma^t R_{t+k}|S_t=s,Z_t=z,A_t=a\bigg].
\end{equation*}
Different from the standard Q-function which is a function of the state-action pair only, our Q-function additionally depends on the IV to handle the unmeasured confounding. 

Based on Theorem \ref{thm:identification}, it is immediate to see that the value function can be represented as a weighted average of the Q-function, i.e., 
\begin{equation}\label{eq:relation_Q_V}
    V^{\pi}(s)=\sum_{z,a}c(z|s)p_a(a|z,s)Q^{\pi}(s,z,a),
\end{equation}
where $p_a(a|z,s):=\mathbb{P}(A_t=a|Z_t=z, S_t=s)$. 
Aggregating \eqref{eq:relation_Q_V} over the empirical initial state distribution yields the DM estimator, which is given by
\begin{equation*}
\widehat{\eta}^{\pi}_{\text{DM}}=\frac{1}{n}\sum_{i,z,a}\widehat{c}(z|S_{i,0})\cdot \widehat{p}_a(a|z,S_{i,0})\widehat{Q}^{\pi}(S_{i,0},z,a),
\end{equation*}
where $\widehat{c}$, $\widehat{p}_a$ and $\widehat{Q}^{\pi}$ denote certain consistent estimators for $c$, $p_a$ and $Q^{\pi}$, respectively. The estimators $\widehat{c}$ and  $\widehat{p}_a$ can be computed via supervised learning, and $\widehat{Q}^{\pi}$ can be obtained by solving a Bellman equation for IV-based MDPUCs. The detailed estimation procedures are summarized in Section \ref{sec:est_details}.

\subsection{Marginal Importance Sampling Estimator}\label{sec:IS}
The second estimator is the marginal importance sampling (MIS) estimator. The traditional stepwise IS estimator, constructed based on the product of individual importance sampling ratios at each time, 
is known to suffer from the curse of horizon \citep{liu2018breaking} and becomes very inefficient in the long-horizon settings. 

To break the curse of horizon, we borrow ideas from \citet{liu2018breaking} and define the marginal importance sampling ratio as below:
\begin{equation*}
    \omega^{\pi}(s)=(1-\gamma)\sum_{t=0}^\infty \gamma^t\frac{p_t^{\pi}(s)}{p_{\infty}(s)},
\end{equation*}
where $p_t^{\pi}$ denotes the probability density/mass function of $S_t$ when the system follows $\pi$, 
and $p_{\infty}(s)$ to denote the stationary distribution of the stochastic process $\{S_t\}_{t\geq 0}$. Thus, it follows from the change of measure theorem that 
\begin{equation*}
    \eta^{\pi}=(1-\gamma)^{-1}\mathbb{E}_{S_t\sim p_\infty} [\omega^{\pi}(S_t)\mathbb{E}^{\pi}(R_t|S_t)]. 
\end{equation*}
By applying the IV-based importance sampling trick detailed in Section 4.2 of \citet{wang2018bounded}, we can represent $\mathbb{E}^{\pi}(R_t|S_t)$ with the observed data distribution and obtain
\begin{equation*} 
\small
\eta^{\pi}=\frac{1}{1-\gamma}\mathbb{E}_{S_t\sim p_\infty}\bigg[  \omega^{\pi}(S_t) \rho(S_t, Z_t) \mathbb{E}[R_t|Z_t,S_t]\bigg],
\end{equation*}
where ${\rho}(s,z)={{c}(z|s)}/{p_z(z|s)}$. 
As such, an MIS estimator can be constructed as below:
\begin{equation}\label{eq:IS}
    \widehat{\eta}_{\text{MIS}}=(1-\gamma)^{-1}\frac{1}{\sum_i T_i}\sum_{i,t}\widehat\omega^{\pi}(S_{i,t})  \widehat{\rho}(S_{i,t},Z_{i,t})R_{i,t},
\end{equation}
where 
$\widehat{\rho}$ and $\widehat\omega^{\pi}$ denote some consistent estimators of ${\rho}$ and $\omega^{\pi}$, respectively. These estimators can be learned from the observed data, as detailed in Section \ref{sec:est_details}.

In Formula \eqref{eq:IS}, the expression for IS estimator consists of two ratios: $\omega^{\pi}(S_{t})$ and ${\rho}(S_t,Z_t)$. 
The second ratio ${\rho}(S_t,Z_t)$ relies on the function $c$ which accounts for the distributional shift, as we have discussed in Remark 2. In the special case where $\pi=b$, we have ${\rho}(s,z)=1$.

Finally, let us conclude this section by briefly discussing the drawbacks of the DM and MIS estimators. Both estimators may be seriously biased due to model misspecifications. Specifically, the consistency of DM requires correct specification of $c$, $p_a$ and $Q^{\pi}$ whereas the consistency of MIS requires correct specification of the two ratio functions. In the next section, we will develop a doubly robust (DR) estimator that combines the strength of both estimators.


\subsection{Our Proposal}\label{sec:EIF}
We begin by deriving the efficient influence function (EIF) for $\eta^{\pi}$, which corresponds to the canonical gradient of a statistical estimand and plays a central role in constructing doubly robust (DR) and semiparametrically efficient estimators \citep{tsiatis2006semiparametric}. The idea of using EIF to develop efficient estimators has been widely used in the statistics and machine learning literature \citep[see e.g.,][]{wang2018bounded,kallus2022efficiently}. 

\begin{theorem}\label{thm:EIF} \textbf{(Efficient Influence Function)}\\
The EIF for $\eta^{\pi}=\mathbb{E}_{S_0\sim \nu}[V^{\pi}(S_0)]$ is given by
\begin{equation}\label{eq:EIF}
    \begin{aligned}    \text{EIF}_{\eta^{\pi}}=&(1-\gamma)^{-1}\omega^{\pi}(S_t)\bigg[\rho(S_t,Z_t)\Big\{Y_t-\mathbb{E}\big[Y_t\big|Z_t,S_t\big]-\left(A_t-\mathbb{E}\left[A_t\big|Z_t,S_t\right]\right)\cdot \Delta(S_t)\Big\}\\
&+\sum_{z_t}c(z_t|S_t)\cdot\mathbb{E}[R_t|z_t,S_t]\bigg]-\eta^{\pi},
    \end{aligned}
\end{equation}
where $\Delta(S_t)$ is defined as the cumulative conditional Wald estimand (cumulative CWE), where
\begin{equation*}
\Delta(S_t)=\frac{\mathbb{E}\left[Y_t\big|Z_t=1,S_t\right]-\mathbb{E}\left[Y_t\big|Z_t=0,S_t\right]}{\mathbb{E}\left[A_t\big|Z_t=1,S_t\right]-\mathbb{E}\left[A_t\big|Z_t=0,S_t\right]},
\end{equation*}
and $Y_t:=R_t+\gamma\cdot V^{\pi}(S_{t+1})$.
\end{theorem}

\noindent\textbf{Remark 4.} The classical CWE plays a key role in identifying the conditional average treatment effect in single-stage decision making. In MDPUCs, we extend the original definition by using $Y_t$ to account for the long-term offline causal effect of executing policy $\pi$. When the discounted factor $\gamma=0$, cumulative CWE will degenerate to the classical CWE.

\noindent\textbf{Remark 5.} We notice that 
a recent concurrent work by \citet{fu2022offline} also developed a DR estimator in IV-based MDPUCs. However, their estimator is not constructed based on the EIF, which is less efficient compared to our proposed DR estimator that will be introduced below.


Based on the result of Theorem \ref{thm:EIF}, we propose a DR estimator $\widehat{\eta}_{\text{DR}}$ for aggregated value $\eta^\pi$, given by
\begin{equation}\label{eq:DR_estimator}  \widehat{\eta}_{\text{DR}}=\widehat{\eta}^{\pi}_{\text{DM}}+(NT)^{-1}\sum_{i,t}\widehat\phi(O_{i,t}),
\end{equation}
where $\widehat{\phi}$ denotes some plug-in estimator for the augmentation function $\phi$:
\begin{eqnarray}\label{eq:psi}
\begin{aligned}
    &\phi(O_t)=(1-\gamma)^{-1}\omega^{\pi}(S_t)\bigg[\rho(S_t,Z_t)\Big\{Y_t-\mathbb{E}\big[Y_t\big|Z_t,S_t\big]-\left(A_t-\mathbb{E}\left[A_t\big|Z_t,S_t\right]\right)\cdot \Delta(S_t)\Big\}.
\end{aligned}
\end{eqnarray}
According to \eqref{eq:DR_estimator}, the proposed estimator is essentially the sum of the DM estimator and an estimated augmentation function $\widehat{\phi}$ which offers additional protection to the final estimator against potential model misspecifications of $Q^{\pi}$. To compute $\widehat{\phi}$, we need to estimate $\omega^{\pi}$, $\rho$, $\mathbb{E}\big[Y_t\big|Z_t,S_t\big]$, $p_a$ and $\Delta$, or equivalently, $\omega^{\pi}$, $p_z$, $p_a$ and $Q^{\pi}$. Since $\mathbb{E}\big[Y_t\big|Z_t,S_t\big]=\sum_{a_t}p_a(a_t|S_t,Z_t)\cdot Q^{\pi}(S_t,Z_t,a_t)$, $\Delta$ and $\rho$ can be determined by $p_z$, $p_a$ and $Q^{\pi}$. 
We will discuss the estimation details of these nuisance functions in Section \ref{sec:est_details}.

Our final estimator $\widehat{\eta}_{\text{DR}}$, as shown in \eqref{eq:DR_estimator}, enjoys the double robustness property.
Firstly, recall that the consistency of $\widehat{\eta}_{\text{DM}}$ relies on the correct specification of $p_a$ and $Q^{\pi}$. When both are correctly specified, so are $\mathbb{E} [Y_t|Z_t,S_t]$ and $\mathbb{E} [A_t|Z_t,S_t]$. As such, it is immediate to see that the augmentation term is mean zero regardless of whether the two IS ratios are correctly specified or not. Therefore, the DR estimator is consistent. 

Secondly, when 
the two IS ratios and $p_a$ are correctly specified, it can be shown that no matter whether $Q^{\pi}$ is correctly specified or not, we have
\begin{equation*}
\begin{aligned}
\mathbb{E}\left[\widehat{\eta}^{\pi}_{\text{DM}}\right]&+(1-\gamma)^{-1}\mathbb{E}\bigg[\omega^{\pi}(S_t)\cdot\rho(S_t,Z_t)\cdot\Big\{\gamma \widehat{V}^{\pi}(S_{t+1})-\sum_{a_t}p_a(a_t|Z_t,S_t) \widehat{Q}^{\pi}(S_t,Z_t,a_t)\Big\}\bigg]=0,
\end{aligned}
\end{equation*} 
where $\widehat{V}^{\pi}$ depends on $\widehat{Q}^{\pi}$ through \eqref{eq:relation_Q_V}. It follows that the DR estimator becomes equivalent to the MIS estimator with correctly specified IS ratios
\[
(NT)^{-1}\sum_{i,t}(1-\gamma)^{-1}\omega^{\pi}(S_{i,t})\rho(S_{i,t},Z_{i,t})\cdot R_{i,t},
\]
and is thus consistent.

\begin{figure}[tbh]
    \centering
    \includegraphics[width=5in]{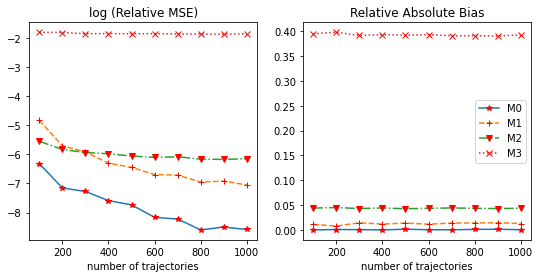}
    \vspace{-10pt}
    \caption{The logarithmic relative MSEs (left panel) and relative absolute biases (right panel) comparison under different model specifications. Specifically, the blue solid line depicts the estimator where the two set of models $\mathcal{M}_1$ and $\mathcal{M}_2$ are correctly specified. The yellow dashed and green dash-dotted lines depict the estimators where one set of the models is correctly specified and the other set misspecified. The red dotted line depicts the estimator where both set of models are misspecified. More details about the data generating process are provided in 
Section \ref{sec:simulation_DR}.
}
    \label{fig:1}
\end{figure}

We empirically verify the doubly robustness property in Figure~\ref{fig:1}. In particular, we apply the proposed method to a toy numerical example detailed in Section \ref{sec:simulation_DR}. It can be seen that the relative absolute bias and MSE of the proposed estimator are fairly small when one set of the models are correctly specified. To the contrary, the resulting estimator is seriously biased when both sets of models are misspecified. 

The following theorem states that $\widehat{\eta}_{\text{DR}}$ is not only doubly robust, but semiparametrically efficient as well (e.g., it achieves the minimum variance or the semiparametric efficiency bound, among all regular and asymptotically linear estimators).
\begin{theorem}\label{thm:DR} 
Suppose that the nuisance function classes  are bounded and belong to VC
type classes \citep{van1996weak} with VC indices upper bounded by $v=O(N^k)$ for some $0\leq k<1/2$.
Define two model classes as below:
\begin{enumerate}
\vspace{-0.5em}
    \item[$\mathcal{M}_1$:]   $Q^{\pi}(s,z,a)$ is correctly specified.
    \vspace{-0.5em}
    \item[$\mathcal{M}_2$:] $p_z(z|s)$ and $\omega^{\pi}(s)$ are correctly specified.
    \vspace{-0.5em}
\end{enumerate}
Suppose $p_a(a|s,z)$ is always correctly specified. Then 

(a) as long as either $\mathcal{M}_1$ or $\mathcal{M}_2$ holds, $\widehat{\eta}_{\text{DR}}$ is a consistent estimator of $\eta^{\pi}$;

(b) when all of the models are correctly specified, and $\widehat{Q}^{\pi}$, $\widehat{p}_a$, $\widehat{p}_z$ and $\widehat{\omega}^{\pi}$ converge in $L_2$ norm (see Appendix \ref{sec:thm3_2} for the detailed definition) to their oracle values at a rate of $o(N^{-\alpha})$ with $\alpha\geq 1/4$,  we have
\[
\sqrt{N}(\widehat{\eta}_{\text{DR}}-\eta^{\pi})\stackrel{d}{\rightarrow} \mathcal{N}(0,\sigma^2_T),
\]
where $\sigma^2_T$ is the efficiency bound of $\eta^{\pi}$, given by
\begin{eqnarray}\label{eq:sigma_T}
    \text{Var}\Big\{ V^{\pi}(S_0)\Big\}+\frac{1}{T^2}\sum_{t=1}^T\text{Var}\Big\{\phi(O_t)\Big\}.
\end{eqnarray}
\end{theorem}

\noindent\textbf{Remark 6. }Theorem \ref{thm:DR}(a) proves the doubly robustness property and (b) proves the semiparametric efficiency. In addition, (b) also establishes the asymptotic normality of $\widehat{\eta}_{\text{DR}}$, based on which 
the following Wald-type confidence interval (CI) can be constructed for $\eta^{\pi}$, 
\[
\Big[\widehat{\eta}_{\text{DR}}\pm z_{\alpha/2}\frac{\widehat{\sigma}_T}{\sqrt{n}}\Big],
\]
where $\widehat{\sigma}^2_T$ is a sampling variance estimator of ${\sigma}^2_T$. 

\noindent\textbf{Remark 7. }It can be seen from \eqref{eq:sigma_T} that the semiparametric efficiency bound $\sigma^2_T$ generally decays with $T$, as we have more data for policy value estimation. In particular, as $T\rightarrow \infty$, the variance of the augmentation term will vanish, resulting the variance bound to be reduced to $\textrm{Var}[V^{\pi}(S_0)]$. 

\subsection{Estimation Details}\label{sec:est_details}
In this section, we summarize the estimation procedures for the models mentioned above. We will first briefly summarize the estimation of some functions that can be easily modeled, and then discuss the estimation of $Q^{\pi}$, $V^{\pi}$ and $\omega^{\pi}$ in the following two subsections.

Estimating $p_z$, $p_a$, and $p_r$ can be treated as standard regression or classification problems, depending on the type of covariates. Any appropriate supervised learning methodology satisfying the convergence rate detailed in Theorem \ref{thm:DR} can be used to estimate these models. Additionally, since $\rho(s_t,z_t)$, $c(z_t|s_t)$ are both functions of $p_z$, $p_a$, $p_r$ and $\pi$, we can first estimate these pdfs/pmfs and then use the resulting estimators to construct plug-in estimators for $\rho$ and $c$.

\subsection{The estimation of $Q^{\pi}$ and $V^{\pi}$}
We first consider the estimation of $Q^{\pi}(s,z,a)$ and $V^{\pi}(s)$. According to Formula (\ref{eq:relation_Q_V}), we can derive the Bellman equation under this confounded MDP as
\begin{equation*}
\small
\begin{aligned}
Q^{\pi}(S_t,Z_t,A_t)=\mathbb{E}\bigg\{R_t+\gamma \sum_{z,a}c(z|S_{t+1})p_a(a|z,S_{t+1}) Q^{\pi}(S_{t+1},z,a)\Big| S_t,Z_t,A_t\bigg\}.
\end{aligned}
\end{equation*}
Motivated by \cite{le2019batch}, we employ fitted-Q evaluation method to iteratively solve the Q function until convergence. Specifically, at the $l$th step, we update $Q^{l+1}$ by
\begin{equation*}
Q^{\pi,l+1}=\arg\min_{Q^{\pi}\in\mathcal{Q}}\sum_{i,t}\Big\{R_{i,t}+\gamma\widehat{V}^{\pi,l}(S_{i,t+1})-Q^{\pi}(S_{i,t},Z_{i,t},A_{i,t})\Big\}^2,
\end{equation*}
where $\mathcal{Q}$ denotes some function class, and $\widehat{V}^{\pi,l}(S_{t+1})=\sum_{z,a}\widehat{c}(z|S_{t+1}) \widehat{p}_a(a|z,S_{t+1}) \widehat{Q}^{\pi,l}(S_{t+1},z,a)$ is the value function calculated from the Q function at the previous step. The algorithm terminates when the maximum number of iterations is reached or a convergence criterion is met. We use the Q function and value function from the final iteration as our estimates of $Q^{\pi}$ and $V^{\pi}$.

\subsection{The estimation of $\omega^{\pi}$}
Then, let's consider the estimation of $\omega^{\pi}(s)$. Define
\[
\begin{aligned}
L(\omega,f)&=\gamma\cdot \mathbb{E}_{(s,a,s')\sim p_t^{\pi}}\left[ \Delta(\omega;s,a,s')\cdot f(s')\right]+(1-\gamma)\cdot \mathbb{E}_{s_0\sim \nu_0(s)}\left[(1-\omega(s))\cdot f(s)\right],
\end{aligned}
\]
where $s'$ denotes the next-state covariates, $\Delta(\omega;s,a,s'):=\omega(s)\cdot \rho(s,z)-\omega(s')$. In confounded MDPs, we can further derive $L(\omega,f)$ as 
\begin{equation}\label{eq:Loss}
\begin{aligned}
&L(\omega,f)
=(1-\gamma)\sum_s f(s)\nu(s)-\mathbb{E}\omega(S_{i,t})\bigg\{f(S_{i,t})-\gamma\cdot \rho(S_{i,t},Z_{i,t})\cdot f(S_{i,t+1})\bigg\}.
\end{aligned}
\end{equation}
According to Theorem 4 in \cite{liu2018breaking}, $\omega^{\pi}(s)$ is the solution to $L(\omega,f)=0$ for any discriminator function $f$. Therefore, ${\omega}^{\pi}$ can be learned by solving the mini-max problem for the quadratic form of the loss function $L(\omega,f)$. Specifically, we aim to find the solution to $\arg\min_{\omega\in \Omega} \sup_{f\in \mathcal{F}} L^2(\omega,f)$
for some function class $\Omega$  and $\mathcal{F}$. 

For the ease of illustrations, let's consider linear bases for $\Omega$ and $\mathcal{F}$. Suppose $\omega^{\pi}(s)=\xi^T(s)\beta$ where $\xi^T(s)$ denotes the basis function. By Formula \eqref{eq:Loss}, $\beta$ can be estimated by
\[
\begin{aligned}
\widehat{\beta}=&\bigg[\sum_{i=1}^N \sum_{t=0}^{T-1}\xi(S_{i,t})\Big\{\xi^T(S_{i,t})-\gamma\widehat{\rho}(S_{i,t},Z_{i,t})\xi^T(S_{i,t+1})\Big\}\bigg]^{-1}\times(1-\gamma)NT\cdot\sum_s \xi(s)\nu(s).
\end{aligned}
\]
Therefore, we can derive the final estimator for 
${\omega}^{\pi}$ as $\widehat{\omega}^{\pi}=\xi^T(s)\cdot \widehat{\beta}$.

\section{Extensions to Non-Markov Settings}\label{sec:extension}
Our proposal in Section \ref{sec:EIF} relies on the set of conditional independence assumptions imposed in Assumption 2. In particular, it requires the states to satisfy the Markov assumption, yielding a memoryless unobserved confounding condition \citep{kallus2020confounding}. This assumption essentially excludes the existence of directed edges from past observed data or $U_{t-1}$ to $U_t$ in Figure \ref{fig:causal_graph} and is likely to be violated in practice. 
In this section, we discuss two potential relaxations of Assumption 2 to accommodate non-Markov settings. Throughout this section, we will use $O_t$ (instead of $S_t$) to denote the time-varying observation measured at time $t$ due to the violation of Markovianity. 

\subsection{High-order MDPs with Unmeasured Confounders}\label{sec:homdpuc}
One approach to relax Markov assumption is to impose a high-order memoryless unobserved confounding condition.  Specifically, a $k$th order memoryless unobserved confounding assumption requires $U_t$ to be conditionally independent of the past data history (including $\{U_j\}_{j<t}$) given $O_t$ and the observation-IV-action triplets collected from time $t-k+1$ to $t-1$. When $k=1$, high-order MDPs will reduce to the memoryless unobserved confounding case. When $k\ge 2$, it allows for the conditional dependence of $U_t$ on the observed data history. 

A key observation is that, under the $k$th order memoryless unobserved confounding assumption, the system forms a $k$th order MDP with unmeasured confounders. Specifically, let $S_t$ denote the union of $O_t$ and the observation-IV-action triplets collected from time $t-k+1$ to $t-1$. By doing so, the newly-defined state satisfies the Markov assumption, i.e., $S_t$ is independent of the past data history given $(S_{t-1},A_{t-1},Z_{t-1})$. As such, our proposal developed in Section \ref{sec:EIF} can be directly applied here to address the $k$th order MDPUC.


\subsection{Partially Observable MDP}\label{sec:POMDP}
To further relax the high-order memoryless assumption, the second approach is to adopt an IV-based POMDP model   
$\mathcal{M}=\langle\mathcal{S},\mathcal{O},\mathcal{Z},\mathcal{A},\mathcal{R}\rangle$ for policy evaluation. Here, $\mathcal{S},\mathcal{O},\mathcal{Z},\mathcal{A},\mathcal{R}$ denote the spaces of latent states, observed features, IVs, actions and rewards, respectively. 
At a given time, suppose the environment is in latent state $S\in \mathcal{S}$. Although $S$ is not directly observable, we have access to an observation $O\sim p_o(\cdot|S)\in \mathcal{O}$. An IV $Z\sim p_z(\cdot|O)$ is generated whose distribution is independent of $S$. Next, based on the action $A\in \mathcal{A}$ of the agent, the environment responds by providing an immediate reward $R$ and transitioning to a new state $S'$. Since $A$, $R$ and $S'$ are all allowed to depend on $S$, the dataset we observed is thus confounded. To proceed, we further denote $H$ and $F$ as the multi-step history and future observations, given by
\begin{eqnarray*}
    H=(O_{t-M_H:t-1},A_{t-M_H:t-1}), \quad F=(O_{t:t+M_F-1},A_{t:t+M_F-2}),
\end{eqnarray*}
where $M_H$ and $M_F$ are two positive integers denoting the number of steps tracing back or forward. As discussed in Section \ref{sec:relatedPOMDP}, several methods have been developed in the literature to handle POMDPs. Here, we extend the proposal developed by \citet{uehara2022future} to IV-based POMDPs to deal with confounders. 

For illustration purpose, we will focus on evaluating memoryless target policies $\pi:\mathcal{O}\rightarrow \mathcal{A}$, but the entire framework can easily be extended to accommodate $M$-memory policies where the decision rule depends on the last $M$ observations. 

In IV-based POMDPs, the Q-function defined in Section \ref{sec:DM} is not directly estimable since the state is not observable. However, due to the temporal dependence, the multi-step history and future observations contain rich information to infer the latent state. These variables serve as proxies for policy value identification. 
Toward that end, we define a 
future-dependent Q-function $g_Q$ 
as the solution to the following conditional moment equation: 
\begin{equation*}
    \begin{aligned}
       \mathbb{E}\Big\{R+\gamma \sum_{z,a}& c(z|O') p_a(a|z,O')g_Q(F',z,a)-g_Q(F,Z,A)\big| H,Z,A\Big\}=0,
    \end{aligned}
\end{equation*}
where $O'$ and $F'$ denote the next-step observation and the next-step future, respectively. Intuitively, $g_Q$ can be viewed as a projection of the Q-function onto the multi-step future. The following theorem shows that the policy value can be consistently identified based on $g_Q$. 

\begin{theorem}\label{thm:POMDP_identification} 
Suppose the following three conditions hold:
\begin{enumerate}
    \item There exists a future-dependent Q function $g_Q$. 
    \item Invertibility: for any $g:\mathcal{S}\times\mathcal{Z}\times\mathcal{A} \rightarrow \mathbb{R}$, if $\mathbb{E}[g(S,Z,A)|H,Z,A]=0$, then $g(S,Z,A)=0$, a.s..
    \item Overlap condition: $|c(Z|O)|<\infty$, a.s..
\end{enumerate}
Then for any $g_Q$, we have
\begin{equation}\label{eq:eta_POMDP}
    \eta^{\pi}=\mathbb{E}_{F\sim \nu_F}\Big[\sum_{z,a} c(z|O)p_a(a|z,O)g_Q(F,z,a)\Big],
\end{equation}
where $\nu_F$ denotes the initial future distribution. 
\end{theorem}

\noindent\textbf{Remark 8. }The first two conditions require the cardinality of the future and the history to be at least greater than or equal to the latent state, respectively. These conditions are weaker than requiring the cardinality of the observation to be greater than or equal to the latent state, which is needed in confounded POMDPs without IVs \citep{nair2021spectral}.

Next, 
we develop a minimax learning approach to estimate $\eta^{\pi}$ from the observed data. According to the result of Theorem \ref{thm:POMDP_identification}, as long as we can learn $g_Q$ from the data, a direct method estimator can be naturally constructed by Equation \eqref{eq:eta_POMDP}.
To address so, we consider the following loss function
\[
\begin{aligned}
\mathcal{L}(q,\xi):= \Big\{R+\gamma \sum_{z,a}c(z| &O') p_a(a|z,O') q(F',z,a)-q(F,Z,A)\Big\}\xi( H,Z,A)
\end{aligned}
\]
for any functions $q$ 
and $\xi$. Given some prespecified function classes $q\in\mathcal{Q}$ and $\xi\in \Xi$, we can solve the following minimax problem to obtain an estimator for $g_Q$:
\begin{eqnarray*}
\widehat{g}_Q=\arg\min_{q\in\mathcal{Q}}\max_{\xi\in \Xi}\mathbb{E}_{\mathcal{D}}\Big[\mathcal{L}(q,\xi)-0.5\lambda\xi^2(H)\Big]
+0.5\alpha'\|q\|^2_{\mathcal{Q}}-0.5\alpha\|\xi\|^2_{\Xi},
\end{eqnarray*}
where $\|\cdot\|^2_{\mathcal{Q}}$ 
and $\|\cdot\|^2_{{\Xi}}$ are certain function norms defined on the spaces of $\mathcal{Q}$ and $\Xi$, and $\lambda$, $\alpha$ and $\alpha'$ are some positive constants.
Closed-form solutions are available when using reproducing kernel Hilbert spaces or linear models to parameterize $\mathcal{Q}$ and $\Xi$ \citep{uehara2020minimax}. 
Given $ \widehat{g}_Q$, $\widehat{c}$ and $\widehat{p}_a$, the resulting DM estimator under POMDP is given by
\begin{equation}\label{eq:POMDP_DM}
\widehat\eta^{\pi}=
\frac{1}{n}\sum_{i=1}^n\Big[\sum_{z,a} \widehat{c}(z|O_{i,0})\widehat{p}_a(a|z,O_{i,0})\widehat{g}_Q(F_{i,0},z,a)\Big]. 
\end{equation}

\subsection{Model Selection}
So far, we have discussed two approaches to relax the memoryless unobserved confounding assumption, one with the high-order memoryless assumption and the other with the POMDP formulation. These assumptions are not directly testable, since they rely on the unmeasured confounders. However, as commented in Section \ref{sec:homdpuc}, under the $k$th order memoryless assumption, the observed data satisfy a $k$th order Markov assumption. When $k=\infty$, this data process becomes a POMDP. This motivates us to apply the sequential testing procedure developed by \citet{shi2020does} for model selection. Specifically, we consider a hypothesis testing problem where 
\[
\begin{aligned}
&H_0: \text{ The system follows an MDP}, \text{ v.s. }\\
&H_1: \text{The system is a high-order MDP or POMDP.}
\end{aligned}
\]
By implementing the forward-backward learning procedure, one can test the $k$th order MDP assumption for any given $k\in\{1,\dots,K\}$. We detail the testing procedure in Algorithm \ref{alg:model_select}.
\begin{algorithm}[tbh]
    \caption{Model Selection for IV-based confounded Off Policy Evaluation}
    \label{alg:model_select}
\begin{algorithmic}
   \State {\bfseries Input:} Data trajectories $\{D_i\}_{1\leq i\leq n}$, parameter $K$.
   \ForAll{$k=1$ {\bfseries to} $K$}
   \State \textbf{Apply} forward-backward learning procedure in Algorithm 1 of \citet{shi2020does}.
   \If{$H_0$ is not rejected}
   \State \textbf{Conclude} the system follows a $k$-th order MDP.
   \State \textbf{Apply} Section \ref{sec:homdpuc} and \eqref{eq:DR_estimator} to estimate $\eta^{\pi}$; \textbf{Break}.
   \EndIf
   \EndFor
   \State \textbf{Conclude} the system is most likely a POMDP.
   \State \textbf{Apply} Section \ref{sec:POMDP} and \eqref{eq:POMDP_DM} to estimate $\eta^{\pi}$.
\end{algorithmic}
\end{algorithm}

\section{Simulation Studies}\label{sec:simulation}

In this section, we will evaluate the performance of our IV-based estimator on synthetic data. We will first use a toy example to demonstrate the double robustness of our estimator, and then conduct detailed comparisons between our estimator and other state-of-the-art methods for OPE estimation under confounded MDPs.

\subsection{Double Robustness}\label{sec:simulation_DR}
\textbf{Data generating process}. For the sake of computational cost, we let $T=100$ and the number of data trajectories $N=\{100,200,\dots,1000\}$. The initial state distribution is generated by a Bernoulli distribution with $p=0.5$, i.e. $S_0\sim \text{Ber}(1,p)$. We define the unmeasured confounder at each stage as $U_t$ as another Bernoulli random variable with $p=0.5$. The instrumental variable $Z_t$, action $A_t$, reward $R_t$ and next state $S_{t+1}$ all follow Bernoulli distributions with the corresponding success rates $\mathbb{P}(Z_t=1)=\text{sigmoid}(S_t+\delta_t-2)$ with $\mathbb{P}(\delta_t=0.25)=\mathbb{P}(\delta_t=0)=0.5$, $\mathbb{P}(A_t=1)=\text{sigmoid}(S_t+2Z_t+0.5U_t-2)$, and $\mathbb{P}(R_t=10)=\mathbb{P}(S_{t+1}=1)=\text{sigmoid}(S_t+A_t+U_t-2)$. In this simulation, we set $U_{t}'=0$ for simplicity, which avoids the confounding between $Z_t$ and $A_t$. However, the confounder between $A_t$ and $(R_t,S_{t+1})$ does exists, which is given by $U_t$. 

In order to evaluate the doubly robust property of our estimator, we use Monte Carlo method to approximate the true models for all functions, and then deliberately introduce shifts that can lead to model misspecification. Specifically, to misspecify $\omega^{\pi}$, we let $\omega^{\pi}_{\text{shifted}}(s_0=1)=\omega^{\pi}_{\text{true}}(s_0=1)/2$, and $\omega^{\pi}_{\text{shifted}}(s_0=0)=2\omega^{\pi}_{\text{true}}(s_0=0)$. To misspecify $p_z$, we define a shift parameter $\alpha\in[0,1]$, and denote $p_{z,\text{shifted}}(z=1|s)=\alpha \cdot p_{z,\text{true}}(z=1|s)+(1-\alpha) \cdot p_{z,\text{true}}(z=0|s)$. To misspecify the Q function $Q^{\pi}$, we define another shift parameter $\beta\in\mathbb{R}$, and let $Q^{\pi}_{\text{shifted}}(s,z,a)=Q^{\pi}_{\text{true}}(s,z,a)+\beta(s,z,a)$. In our simulation setup, we fix $\alpha=0.55$, and set $\beta(s,z,a)\sim\mathcal{N}(5,4)$.

\noindent\textbf{Results}. The results are shown in Figure \ref{fig:1}. The comparison of MSEs and biases demonstrate that the performance of $\mathcal{M}_3$ is significantly worse than that of $\mathcal{M}_0$, $\mathcal{M}_1$, and $\mathcal{M}_2$, supporting the consistency of our estimator when at least one group of the models in Theorem \ref{thm:DR} is correctly specified. Moreover, as the number of trajectories increases, the MSEs for $\mathcal{M}_0$, $\mathcal{M}_1$, and $\mathcal{M}_2$ decrease towards zero. When all models are correctly specified, the blue line yields the best performance, demonstrating the efficiency (Theorem \ref{thm:EIF}) of our approach.

\subsection{Comparison With Other Approaches}
In this section, we compare the proposed estimator in Section \ref{sec:EIF} (denoted by IVMDP) against several baseline methods that ignore the unmeasured confounding.

\noindent\textbf{Data generating process}. The observed data consists of $N=1000$ trajectories, each with $T=100$ time points. We consider a two-dimensional state variable $S_{t}=(S_{t,1},S_{t,2})$ whose initial distribution is given by $\mathcal{N}(\boldsymbol{0}_2,I_2)$ where $I_2$ denotes a two-dimensional identity matrix. The unmeasured confounders $\{U_t\}_t$ follow i.i.d. Rademacher distributions. Both the IV and the action are binary. At each time, they satisfy $\mathbb{P}(Z_t=1|S_t)=\text{sigmoid}(S_{t,1}+S_{t,2})$ and $\mathbb{P}(A_t=1|S_t,Z_t,U_t)=\text{sigmoid}\{S_{t,1}+S_{t,2}+2Z_t+U_t\}$, respectively. 
Finally, the reward and next-state are generated as follows: $R_t=S_{t,1}+S_{t,2}+2A_t+2.5U_t$, 
$S_{t+1,1}=S_{t1}+0.5U_t+A_t-0.5$, $S_{t+1,2}=S_{t1}-0.5U_t-A_t+0.5$. 

\noindent\textbf{Competing methods}. We consider three baseline methods, corresponding to the DM estimator, the MIS estimator \citep{liu2018breaking} and the DRL estimator \citep{kallus2022efficiently}. All the estimators are derived under the NUC assumption without the use of IV, denoted by NUC-DM, NUC-MIS and NUC-DRL, respectively. To ensure a fair comparison, we also incorporate the IV in the state variable when implementing the three baseline approaches.

\begin{figure}[tbh]
    \centering
    \includegraphics[width=5in]{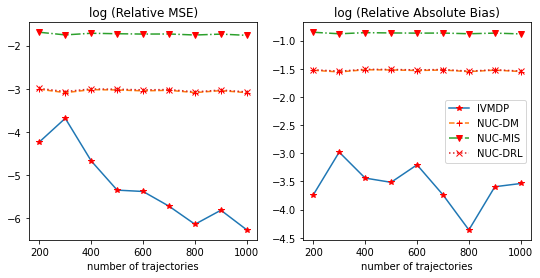}
    \vspace{-10pt}
    \caption{Logarithmic relative MSE (left panel) and logarithmic relative absolute bias (right panel) of various estimators with sample size on the x-axis. Notice that the yellow dashed line and the red dotted line are largely overlapped due to the similar performance under NUC-DM and NUC-DRL.}
    \label{fig:sec5.2}
\end{figure}

The first competing method is a direct estimator (NUC-DM), which is represented by the yellow dashed line in Figure \ref{fig:sec5.2}. When NUC assumption holds, the Bellman equation becomes 
\[
    \mathbb{E}\bigg\{R_t+\gamma \cdot\sum_{a}\pi(a|S_{t+1})\cdot Q^{\pi}(S_{t+1},a)\Big| S_t,A_t\bigg\}=Q^{\pi}(S_t,A_t).
\]
Thus, we can conduct fitted Q evaluation to repeatedly estimate $Q^{\pi}(s,a)$ and $V^{\pi}(s)$ until convergence:
$$ Q^{\pi,l+1}=\arg\min_{Q^{\pi}\in\mathcal{Q}}\sum_{i,t}\bigg\{R_{i,t}+\gamma\widehat{V}^{\pi,l}(S_{i,t+1})-Q^{\pi}(S_{i,t},A_{i,t})\bigg\}^2,
$$
where $\mathcal{Q}$ is some function class, and $\widehat{V}^{\pi,l}(S_{t+1})=\sum_{a}\pi(a|S_{t+1})\cdot \widehat{Q}^{\pi,l}(S_{t+1},a)$ is the value function calculated from the Q function at the $l$th step.
As such, the final NUC-DM estimator is given by 
\[
\widehat{\eta}_{\text{NUC-DM}}^{\pi}=\sum_{a,s_0}\pi(a|s_0)\cdot \widehat{Q}^{\pi}(s_0,a)\cdot \nu(s_0).
\]

The second estimator is an MIS estimator (NUC-MIS), which is represented by the green dash-dotted line in Figure \ref{fig:sec5.2}. According to \cite{liu2018breaking}, we can calculate the NUC-MIS estimator by
\[
\widehat{\eta}_{\text{NUC-MIS}}^{\pi}=(1-\gamma)^{-1}\frac{1}{\sum_i T_i}\sum_{i,t}\hat\omega^{\pi}(S_{i,t})\cdot  \hat{\beta}_{\pi/\pi_0}(a,s)\cdot R_{i,t},
\]
where $\beta_{\pi/\pi_0}(a,s)={\pi(a|s)}/{\pi_0(a|s)}$ and $\hat\omega^{\pi}(S_t)$ can be obtained from the method provided in the original paper. Details are omitted here.

The third estimator is the double reinforcement learning estimator (NUC-DRL), which is represented by the red dotted line in Figure \ref{fig:sec5.2}. DRL combines the NUC-DM and NUC-MIS estimators to provide a more robust estimator under the NUC assumption \citep{kallus2022efficiently}. The final estimator is given by
\[
\begin{aligned}
&\widehat{\eta}_{\text{NUC-DRL}}^{\pi}=\widehat{\eta}_{\text{NUC-DM}}^{\pi}+\widehat{\phi}_{\text{NUC-aug}}=\sum_{a,s_0}\pi(a|s_0)\cdot \widehat{Q}^{\pi,l}(s_0,a)\cdot \nu(s_0)+\\
&(1-\gamma)^{-1}\frac{1}{\sum_i T_i}\sum_{i,t}\hat\omega^{\pi}(S_{i,t}) \hat{\beta}_{\pi/\pi_0}(a,s) \big\{R_{i,t}+\gamma\widehat{V}^{\pi}(S_{i,t+1})-\widehat{Q}^{\pi}(S_{i,t},A_{i,t})\big\}.
\end{aligned}
\]

\noindent\textbf{Results}. The results are shown in Figure \ref{fig:sec5.2}. We can see that our proposed estimator IVMDP achieves the smallest MSE and bias in all cases. Its MSE generally decays with an increase in the number of trajectories, demonstrating the consistency of our proposal. In contrast, other estimators are severely biased, highlighting the risk of ignoring unobserved confounding. The biases of baseline methods dominate the standard deviations, resulting in the MSEs to be relatively constant despite the increase in the number of trajectories.

\section{Real Data Analysis}\label{sec:realdata}
In this section, we apply our method to a real dataset from a world-leading technological company. 
The company conducts advertising compaigns to attract consumers to download their mobile app products. 
The advertisements are delivered through multiple media channels, such as search, display, social, mobile and video, provided by ads exchange or mobile application stores. During the compaign, an individual user is typically exposed to various advertisements 
delivered through these channels. 
To improve the return on investment, it is crucial for the company to 
accurately evaluate the long-term effects of different ads exposure policies. 

The dataset is collected from a randomized advertising campaign. At each time, the company randomly decided whether to bid against other firms or not to display their ad to a target consumer. As such, the IV independence assumption (see Assumption 1(a)) is automatically satisfied. In addition, if the company chooses to bid, it will largely increase the chance that their ad is indeed displayed to the consumer. This meets the IV relevance assumption (see Assumption 1(b)). Finally, bidding can only affect the conversion rate or the consumer behavior through the ad exposure. This verifies the 
exclusion restriction assumption (see Assumption 1(c) \& (e)). The core IV assumptions are thus satisfied in our example. 


Due to privacy considerations, we generate a synthetic data environment based on the real data and report the performance of our proposal applied to this environment. Specifically, we adopt the IV-based MDPUC model to model the data generating process and leverage the IV to estimate the reward and next-state distributions in the presence of unmeasured confounding. 
\begin{figure}[tbh]      
    \centering
    \includegraphics[width=6in]{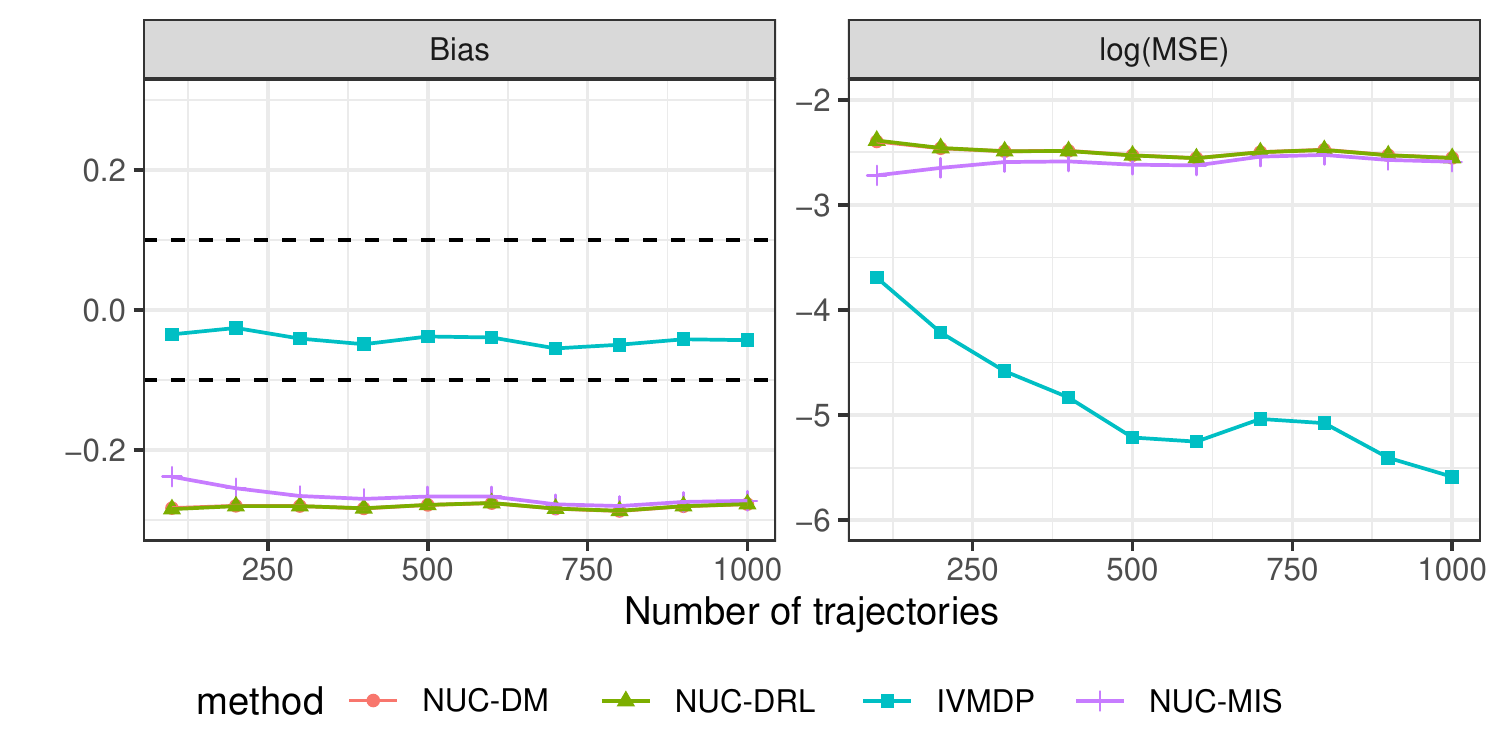}
    \vspace{-24pt}
    \caption{Bias (left panel) and logarithmic MSE (right panel) and  as the sample size increases. The two dashed lines correspond to the Monte Carol error bounds for the bias. As we can see from the figure, only our approach is empirically unbiased and achieves the smallest MSE.}
    \label{fig:real-data}
\end{figure}
The conditional distribution of reward $R_t$ given $S_t, Z_t, A_t$ is modeled by a logistic regression, i.e. $\hat{p}_r(R_t=1|S_t,Z_t,A_t) = \sigma((1,S_t^T, Z_t, A_t)^T\beta_r)$ where $\sigma(x) = \frac{1}{1+\exp(-x)}$ is the sigmoid function. Similarly, we estimate $\mathbb{E}(S_{t+1} | S_t, Z_t, A_t)$ by fitting a multivariate linear model $T(S_t, Z_t, A_t)$ with response $\{S_{i, t+1}\}$ and covariates $\{(S_{i, t}, Z_{i, t}, A_{i, t})\}$. Then, the transition model of $S_{t+1} | S_t, Z_t, A_t$ is characterized by $N(T(S_t, Z_t, A_t), \textup{diag}\{\sigma_1^2, \ldots, \sigma_6^2\})$ where $\sigma^2_i (i=1, \ldots, 6)$ are estimated by the residual of the linear model for $S_i$. For the transition from $S_t, Z_t$ to $A_t$, we model it with a logistic regression, which is denoted as $\hat{p}_a(A_t=1|S_t, Z_t) = \sigma((1, S_t^T, Z_t)^T \beta_a)$. Finally, since $Z_t$ is independent to $S_t$, we simply model it by a binomial random variable with probability $p$ in which $p$ is estimated by empirical frequency.

We depict the procedure for computing the true policy value of target policy: $\pi(1|S_t) = \sigma((1, S_t^T, \mathbb{E}(Z_t))^\top \beta)$. 
Motivated by our identification result in Theorem~\ref{thm:identification}, we can use Monte Carlo method to estimate $\eta^\pi$. More precisely, we draw $N'$ trajectories with length $T$ from the fitted MDP and attain the transition tuple $\{S_{i, t}, Z_{i, t}, A_{i, t}, R_{i, t}\}_{i=1, t=1}^{N', T}$. Therefore, we compute 
$$\eta_{\textup{MC}} = \frac{1}{N'}\sum_{i=1}^{N'}\sum_{t=j}^{T} R_{i, t} \prod_{j=0}^t\frac{c(Z_{i, j}|Z_{i, j})}{p_z(Z_{i, j}|S_{i, j})}$$ 
as the estimator for $\eta^\pi$. To guarantees the accuracy of $\eta_{\textup{MC}}$, we set $N'=10^8$ to overcome the large variance of $\prod_{j=0}^t\frac{c(Z_{i, j}|S_{i, j})}{p_z(Z_{i, j}|S_{i, j})}$.

The numerical results are reported in Figure~\ref{fig:real-data}. First, we can see that 
our proposed estimator achieves the least bias and MSE in all cases. In particular, the bias of IVMDP remains close to zero, and its MSE decreases with an increase in the number of trajectories. 
In contrast, other estimators significantly underestimate the true value, and their MSEs do not decay with the number of trajectories. This demonstrates that our estimator is able to effectively handle unmeasured confounders, while other estimators are considerably biased.

\section{Future Work}
\label{sec:conc}

In this paper, we presented a systematic approach for using instrumental variables to perform off-policy evaluation in infinite-horizon confounded MDPs. To the best of our knowledge, this is the first work to derive the efficient influence function of the value function in an IV-based confounded MDP, high-order MDP and POMDP. Our numerical results in simulation and real data analysis both demonstrate the effectiveness of our method.

There are several potential avenues for future work that could build on the advances presented in this paper. One possibility is to extend the framework to handle discrete or continuous action spaces, as has been done in previous work such as \citet{heckman2008instrumental,cai2020deep}. Another option is to further explore more efficient estimators, such as IS, DR estimators under confounded POMDP by continuing our discussion in Section \ref{sec:extension}.

\bibliographystyle{chicago}
\bibliography{Reference}


\clearpage
\appendix
\begin{center}
{\large\bf SUPPLEMENTARY MATERIAL}
\end{center}

\section{Proof of Theorem \ref{thm:identification}}\label{appendix:ident}
Theorem 1 states the identifiability of the value function, i.e. $V^{\pi}(s_0)$ can be entirely estimated from the observed data. 

Since the value function is defined as $V^{\pi}(s_0)=\sum_{t=0}^{\infty}\gamma^t\mathbb{E}^{\pi}\left[R_t|S_0=s_0\right]$,
it suffice to identify $\mathbb{E}^{\pi}\left[R_t|S_0=s_0\right]$ under each stage $t$. This expression, under Assumption 2, can be further decomposed as
\begin{equation}\label{eq:ident_1}
\begin{aligned}
\mathbb{E}^{\pi}\left[R_t|S_0=s_0\right]&=\mathbb{E}\left[R_t(\pi)|S_t=s_t\right]\cdot \mathbb{P}(S_t(\pi)|S_{t-1}=s_{t-1}) \cdots \mathbb{P}(S_1(\pi)|S_{0}=s_0),
\end{aligned}    
\end{equation}
where $R_t(\pi)$ is defined as the potential reward one would observe under target policy $\pi$. 

Before we proceed, let's define some short-hand notations that will be widely used in the following sections.
First, we define
\[
p_{1}^{A}(S_t)=\mathbb{E}[A_t|Z_t=1,S_t],\quad p_{0}^{A}(S_t)=\mathbb{E}[A_t|Z_t=0,S_t],\quad \delta^{A}(S_t)=p_1^{A}(S_t)-p_0^{A}(S_t),
\]
where the subscript in $p_{1}^{A}(S_t)$ and $p_{0}^{A}(S_t)$ indicates the value of $Z_t$ that $A_t$ is conditioning on. Let $Y_t=R_t+\gamma V^{\pi}(S_{t+1})$, which accounts for the reward at current stage and beyond. We'll see in later sections that instead of using $R_t$, $Y_t$ will be frequently used in our identification and estimation process. Akin to $A_t$, one can similarly define
\[
\begin{aligned}
p_{1}^{Y}(S_t)=\mathbb{E}[Y_t &|Z_t=1,S_t],\quad p_{0}^{Y}(S_t)=\mathbb{E}[Y_t|Z_t=0,S_t],\quad 
\delta^{Y}(S_t)=p_{1}^{Y}(S_t)-p_{0}^{Y}(S_t).\\
&p_{1}^{S}(S_t)=\mathbb{E}[S_{t+1}|Z_t=1,S_t],\quad p_{0}^{S}(S_t)=\mathbb{E}[S_{t+1}|Z_t=0,S_t].
\end{aligned}
\]
Finally, we let $p_r(r_t|a_t,S_t)=\mathbb{P}_r(R_t=r_t|A_t=a_t,S_t)$, $p_a(a_t|z_t,S_t)=\mathbb{P}_a(A_t=a_t|Z_t=z_t,S_t)$, and $p_z(z_t|S_t)=\mathbb{P}(Z_t=z_t|S_t)$ denote the probability density function for $R_t$, $A_t$ and $Z_t$ respectively.

According to \eqref{eq:ident_1}, the identification procedure of $V^{\pi}(s_0)$ can be conducted stage-by-stage. In the following steps, we will first focus on the identifiability of each term on the right hand side of Formula \eqref{eq:ident_1}, and summarize our identification results in Step 3.

\noindent\textbf{Step 1.} Identifiability of $\mathbb{E}\left[R_t(\pi)|S_t\right]$.
    
    First, suppose $a\in [0,1]$ is a constant, and we define $C(z):=2az+1-z-a$ as a function of $z$. For the simplicity of notations, we will drop $S_t=s_t$ whenever calculating a conditional expectation, and omit $\bar{a}_{t-1}$ in $R_t(1,\bar{a}_{t-1})$ and $S_{t+1}(1,\bar{a}_{t-1})$.
    \[
    \begin{aligned}
    &a\mathbb{E}[R_t|Z_t=1]+(1-a)\mathbb{E}[R_t|Z_t=0]\\&=\sum_{z_t=0,1}C(z_t)\cdot \mathbb{E}[R_t|Z_t=z_t]\stackrel{A1(a)}{\xlongequal[]{}}\sum_{z_t=0,1}C(z_t)\mathbb{E}_{U_t}\big[\mathbb{E}[R_t|Z_t=z_t,U_t]\big]\\&=\sum_{z=0,1}C(z_t)\mathbb{E}_{U_t}\Big[\mathbb{E}[R_tA_t|Z_t=z_t,U_t]+\mathbb{E}[R_t(1-A_t)|Z_t=z_t,U_t]\Big]\\
    &\stackrel{A1(d)}{\xlongequal[]{}}\mathbb{E}_{U_t}\sum_{z_t=0,1}C(z_t)\Big\{\mathbb{E}[R_t(1)|Z_t=z_t,U_t]\mathbb{P}(A_t=1|Z_t=z_t,U_t)\\
    &\qquad\qquad\qquad\qquad\quad+\mathbb{E}[R_t(0)|Z_t=z_t,U_t]\mathbb{P}(A_t=0|Z_t=z_t,U_t)\Big\}\\
    &\stackrel{A1(d)}{\xlongequal[]{}}\mathbb{E}_{U_t}\sum_{z_t=0,1}C(z_t)\Big\{\mathbb{E}[R_t(1)|U_t]\mathbb{P}(A_t=1|Z_t=z_t,U_t)+\mathbb{E}[R_t(0)|U_t]\mathbb{P}(A_t=0|Z_t=z_t,U_t)\Big\}\\
    &=\mathbb{E}_{U_t}\Big\{\mathbb{E}[R_t(0)|U_t]+(\mathbb{E}[R_t(1)-R_t(0)|U_t])(a\mathbb{E}[A_t|Z_t=1,U_t]+(1-a)\mathbb{E}[A_t|Z_t=0,U_t])\Big\}\\
    &\stackrel{A1 (g)}{\xlongequal[]{}}\mathbb{E}[R_t(0)]+\mathbb{E}[R_t(1)-R_t(0)]\cdot \left\{a\mathbb{E}[A_t|Z_t=1]+(1-a)\mathbb{E}[A_t|Z_t=0]\right\}.
    \end{aligned}
    \]
    When $a=1$, we have 
    \begin{equation}\label{eq:a=1}
    \begin{aligned}
    &\mathbb{E}[R_t|Z_t=1]=\mathbb{E}_{U_t}\Big\{\mathbb{P}(A_t=0|Z_t=1,U_t)\cdot\mathbb{E}[R_t(0)|U_t]+\mathbb{P}(A_t=1|Z_t=1,U_t)\cdot\mathbb{E}[R_t(1)|U_t]\Big\}\\
    &\stackrel{A1 (g)}{\xlongequal[]{}}\mathbb{E}_{U_t}\Big\{\mathbb{P}(A_t=0|Z_t=1)\cdot\mathbb{E}[R_t(0)|U_t]+\mathbb{P}(A_t=1|Z_t=1)\cdot\mathbb{E}[R_t(1)|U_t]\Big\}\\
    &=\mathbb{P}(A_t=0|Z_t=1)\cdot\mathbb{E}[R_t(0)]+\mathbb{P}(A_t=1|Z_t=1)\cdot\mathbb{E}[R_t(1)]\end{aligned}      
    \end{equation}
    Similarly, when $a=0$,
    \begin{equation}\label{eq:a=0}
    \begin{aligned}
    &\mathbb{E}[R_t|Z_t=0]=\mathbb{E}_{U_t}\Big\{\mathbb{P}(A_t=0|Z_t=0,U_t)\cdot\mathbb{E}[R_t(0)|U_t]+\mathbb{P}(A_t=1|Z_t=0,U_t)\cdot\mathbb{E}[R_t(1)|U_t]\Big\}\\
    &\stackrel{A1 (g)}{\xlongequal[]{}}\mathbb{E}_{U_t}\Big\{\mathbb{P}(A_t=0|Z_t=0)\cdot\mathbb{E}[R_t(0)|U_t]+\mathbb{P}(A_t=1|Z_t=0)\cdot\mathbb{E}[R_t(1)|U_t]\Big\}\\
    &=\mathbb{P}(A_t=0|Z_t=0)\cdot\mathbb{E}[R_t(0)]+\mathbb{P}(A_t=1|Z_t=0)\cdot\mathbb{E}[R_t(1)]\end{aligned}
    \end{equation}
    By solving for Equation \eqref{eq:a=1} and \eqref{eq:a=0}, one can obtain that
    \[
    \begin{aligned}
    \mathbb{E}[R_t(0)|S_t]&=\frac{\mathbb{E}[R_t|Z_t=0,S_t]\cdot \mathbb{P}(A_t=1|Z_t=1,S_t)-\mathbb{E}[R_t|Z_t=1,S_t]\cdot \mathbb{P}(A_t=1|Z_t=0,S_t)}{\mathbb{P}(A_t=1|Z_t=1,S_t)-\mathbb{P}(A_t=1|Z_t=0,S_t)},\\ 
    \mathbb{E}[R_t(1)|S_t]&=\frac{\mathbb{E}[R_t|Z_t=1,S_t]\cdot \mathbb{P}(A_t=0|Z_t=0,S_t)-\mathbb{E}[R_t|Z_t=0,S_t]\cdot \mathbb{P}(A_t=0|Z_t=1,S_t)}{\mathbb{P}(A_t=1|Z_t=1,S_t)-\mathbb{P}(A_t=1|Z_t=0,S_t)}.
    \end{aligned}
    \]
    Notice that all of the terms on the RHS of the expression above can be estimated from the observed data. Specifically, $\mathbb{E}[R_t|Z=1,S_t]$ can be further written as
    \[
    \begin{aligned}\mathbb{E}[R_t|Z_t=1,S_t]&=\sum_{a_t,r_t}r_t\cdot\mathbb{P}_r(R_t=r_t|Z_t=1,A_t=a_t,S_t)\cdot \mathbb{P}_a(A_t=a_t|Z_t=1,S_t)
    \end{aligned}
    \]
    which is also estimable.
    To sum up, the identification result of $\mathbb{E}\left[R_t(\pi)|S_t\right]$ is given by
    \begin{equation}\label{eq:ident_2}
    \begin{aligned}
    &\mathbb{E}\left[R_t(\pi)|S_t\right]=\pi(1|S_t)\cdot \mathbb{E}\left[R_t(1)|S_t\right]+(1-\pi(1|S_t))\cdot \mathbb{E}\left[R_t(0)|S_t\right]\\&=\frac{\mathbb{E}[R_t|Z_t=0,S_t]\cdot \{p_{1}^{A}(S_{t})-\pi(1|S_t)\}-\mathbb{E}[R_t|Z_t=1,S_t]\cdot \{p_{0}^{A}(S_{t})-\pi(1|S_t)\}}{\delta^{A_{t}}(S_{t})}.
    \end{aligned}
    \end{equation}
    Therefore, $\mathbb{E}\left[R_t(\pi_t)|S_t\right]$ is identifiable.

\vspace{0.1in}
\noindent\textbf{Step 2: }Identifiability of $\mathbb{P}(S_t(\pi)|S_{t-1}=s_{t-1})$.

Under the potential outcome's framework, it holds for $\mathbb{P}(S_t(\pi)|S_{t-1})$ that
    \[
    \mathbb{P}(S_t(\pi)|S_{t-1})=\pi(S_{t-1})\cdot \mathbb{P}(S_t(1)|S_{t-1})+(1-\pi(S_{t-1}))\cdot \mathbb{P}(S_t(0)|S_{t-1}).
    \]
    In order to indentify $\mathbb{P}(S_t(\pi)|S_{t-1})$, it suffice to derive the identification result for both $\mathbb{P}(S_t(1)|S_{t-1})$ and $\mathbb{P}(S_t(0)|S_{t-1})$.
    Akin to what we did in Step 1, under Assumption 1 (a), (b), (f) and (g), we can show that
    \[
    \begin{aligned}
    &\mathbb{P}[S_t(0)|S_{t-1}]=\frac{\mathbb{P}[S_t|Z_{t-1}=0,S_{t-1}]\cdot p_{1}^{A_{t}}(S_{t-1})-\mathbb{P}[S_t|Z_{t-1}=1,S_{t-1}]\cdot p_{0}^{A_{t}}(S_{t-1})}{\delta^{A_{t}}(S_{t-1})}\\ &\mathbb{P}[S_t(1)|S_{t-1}]=\frac{\mathbb{P}[S_t|Z_{t-1}=1,S_{t-1}]\cdot (1-p_{0}^{A_{t}}(S_{t-1}))-\mathbb{P}[S_t|Z_{t-1}=0,S_{t-1}]\cdot (1-p_{1}^{A_{t}}(S_{t-1}))}{\delta^{A_{t}}(S_{t-1})}.
    \end{aligned}
    \]
    Thus, the identification result of $\mathbb{P}(S_t(\pi)|S_{t-1}=s_{t-1})$ is given by
    \begin{equation}\label{eq:ident_4}
        \begin{aligned}\mathbb{P}&(S_t(\pi)|S_{t-1}=s_{t-1})=\pi(S_{t-1})\cdot\mathbb{E}\left[S_t(1)|S_{t-1}\right]+(1-\pi(S_{t-1}))\cdot\left[S_t(0)|S_{t-1}\right]\\&=\frac{p_0^{S}(s_{t-1})\cdot \{p_{1}^{A}(S_{t-1})-\pi(S_{t-1})\}-p_1^{S}(s_{t-1})\cdot \{p_{0}^{A}(S_{t-1})-\pi(S_{t-1})\}}{\delta^{A}(S_{t-1})}.\end{aligned}    
    \end{equation}

\noindent \textbf{Step 3: } By repeating the procedure in Step 2, it's easy to show that all of the terms on the RHS of \eqref{eq:ident_1} are identifiable.

Define a weighted function $c(z|s)$ as
\begin{equation*}
c(z_t|S_t)=\left\{\begin{aligned}
 \frac{\mathbb{E}[A_t|Z_t=1,S_t]-\pi(1|S_t)}{\mathbb{E}[A_t|Z_t=1,S_t]-\mathbb{E}[A_t|Z_t=0,S_t]}, &\quad\text{when } z_t=0 \\
\frac{\pi(1|S_t)-\mathbb{E}[A_t|Z_t=0,S_t]}{\mathbb{E}[A_t|Z_t=1,S_t]-\mathbb{E}[A_t|Z_t=0,S_t]},  &\quad\text{when } z_t=1
\end{aligned}\right..
\end{equation*}
Therefore, we can rewrite (\ref{eq:ident_2}) and \eqref{eq:ident_4} as
\[
\begin{aligned}
\mathbb{E}\left[R_t(\pi)|S_t\right]&=\sum_{a_t,r_t}r_t\cdot\Big\{c(Z_t=0|S_t)  \cdot p_r(r_t|a_t,Z_t=0,S_t)\cdot p_a(a_t|Z_t=0,S_t)\\
 &\qquad \qquad +c(Z_t=1|S_t)\cdot p_r(r_t|a_t,Z_t=1,S_t)\cdot p_a(a_t|Z_t=1,S_t)\Big\}\\
&=\sum_{z_t,a_t,r_t}r_t\cdot p_r(r_t|a_t,z_t,S_t)\cdot p_a(a_t|z_t,S_t)\cdot c(z_t|S_t)\\
&=\sum_{z_t,a_t,r_t,s_{t+1}}r_t\cdot p_{r,s}(r_t,s_{t+1}|a_t,z_t,S_t)\cdot p_a(a_t|z_t,S_t)\cdot c(z_t|S_t),
\end{aligned}
\]
and similarly,
\[
\begin{aligned}
\mathbb{E}\left[S_t(\pi)|S_{t-1}\right]&=\sum_{a_{t-1},s_{t}}\Big\{c(Z_{t-1}=0|S_{t-1})  \cdot p_s(s_t|a_{t-1},Z_{t-1}=0,S_{t-1})\cdot p_a(a_{t-1}|Z_{t-1}=0,S_{t-1})\\
 &\qquad \qquad +c(Z_{t-1}=1|S_{t-1})\cdot p_s(s_t|a_{t-1},Z_{t-1}=1,S_{t-1})\cdot p_a(a_{t-1}|Z_{t-1}=1,S_{t-1})\Big\}\\
&=\sum_{z_{t-1},a_{t-1},s_t} p_s(s_t|a_{t-1},z_{t-1},S_{t-1})\cdot p_a(a_{t-1}|z_{t-1},S_{t-1})\cdot c(z_{t-1}|S_{t-1})\\
&=\sum_{z_{t-1},a_{t-1},r_{t-1},s_{t}} p_{r,s}(r_{t-1},s_{t}|a_{t-1},z_{t-1},S_{t-1})\cdot p_a(a_{t-1}|z_{t-1},S_{t-1})\cdot c(z_{t-1}|S_{t-1}).
\end{aligned}
\]
Repeating this process until $t=1$, we have
\[
\begin{aligned}
\mathbb{E}^{\pi}\left[R_t|S_0=s_0\right]&=\mathbb{E}\left[R_t(\pi)|S_t=s_t\right]\cdot \mathbb{P}(S_t(\pi)|S_{t-1}=s_{t-1}) \cdots \mathbb{P}(S_1(\pi)|S_{0}=s_0)\\
&=\sum_{\{z_j,a_j,r_j,s_{j+1}\}_{j=0}^t}r_t\cdot\bigg\{ \prod_{j=0}^t p_{r,s}(r_j,s_{j+1}|a_j,z_j,s_j)\cdot p_a(a_j|z_j,s_j)\cdot c(z_j|s_j)\bigg\}
\end{aligned}
\]
    Therefore, the value function $V^{\pi}(s_0)$ can be written as
    \[
    \begin{aligned}
    V^{\pi}(s_0)&=\sum_{t=0}^{\infty}\gamma^t\mathbb{E}^{\pi}\left[R_t|S_0=s_0\right]\\
    &=\sum_{t=0}^{T}\sum_{\{z_j,a_j,r_j,s_{j+1}\}_{j=0}^t}\gamma^tr_t\cdot\bigg\{ \prod_{j=0}^t p_{r,s}(r_j,s_{j+1}|a_j,z_j,s_j)\cdot p_a(a_j|z_j,s_j)\cdot c(z_j|s_j)\bigg\},
    \end{aligned}
    \]
    where the RHS is purely constructed from the observed data. 
Furthermore, the identification result of $\eta^{\pi}$ can be obtained by taking the expectation of $V^{\pi}(s_0)$ w.r.t. the initial state distribution $\nu(s_0)$, which is given by
\begin{equation}\label{eq:identification_eta}
\small
\begin{aligned} &\eta^{\pi}=\sum_{s_0}\nu({s_0})\cdot\Bigg[\sum_{t=0}^{T}\sum_{\{z_j,a_j,r_j,s_{j+1}\}_{j=0}^t}\gamma^tr_t\cdot\bigg\{ \prod_{j=0}^t p_{r,s}(r_j,s_{j+1}|a_j,z_j,s_j)\cdot p_a(a_j|z_j,s_j)\cdot c(z_j|s_j)\bigg\}\Bigg].   
\end{aligned}
\end{equation}
Therefore, $V^{\pi}(s_0)$ and $\eta^{\pi}$ are identifiable.

\section{Proof of Theorem \ref{thm:EIF}}
Define $\mathcal{M}$ as the collection of all models involved in estimating $p_{s}$, $p_r$, $p_a$, $p_z$, $Q^{\pi}$, $\omega^{\pi}$ and $\nu$. Specifically, we suppose that there exists a parameter $\theta_0$ such that $p_{s,
\theta_0}$, $p_{r,\theta_0}$, $p_{a,\theta_0}$, $p_{z,\theta_0}$, $Q^{\pi}_{\theta_0}$, $\omega^{\pi}_{\theta_0}$ and $\nu_{\theta_0}$ correspond to the true models. As such, the aggregated value $\eta_{\pi}$ can also be written as a function of $\theta$, where we denote it as $\eta^{\pi}_{\theta}$. Furthermore, we define $p_D(s)=\frac{1}{T}\sum_{t=0}^{T-1}p_t^b(s)$ as the mixture distribution of the observed data. When the stochastic process is stationary, $p_D(s)=p_{\infty}(s)$. 
 To find the efficient influence function (EIF) for $\eta^{\pi}$, we first need to find the canonical gradient in the nonparametric model $\mathcal{M}$.

\noindent \textbf{Step 1.} By definition, the Cramer-Rao Lower Bound is
$$
CR(\mathcal{M})=\left(\frac{\partial\eta_{\theta}^{\pi}}{\partial \theta_0}\right)\left(\mathbb{E}\left[\frac{\partial l(\{O_t\};\theta)}{\partial \theta_0}\frac{\partial^T l(\{O_t\};\theta)}{\partial \theta_0}\right]\right)^{-1}\left(\frac{\partial\eta_{\theta}^{\pi}}{\partial \theta_0}\right)^T,
$$
where $l(\{O_t\};\theta)$ denotes the log-likelihood function of the observed data, which can be expressed as
$$
l(\{O_t\};\theta)=\sum_{t=0}^{T-1} \log p_{s,r,\theta}(S_{t+1},R_j|A_t,Z_t,S_t)+\sum_{t=0}^{T-1} \log p_{a,\theta}(A_t|Z_t,S_t)+\sum_{t=0}^{T-1} \log p_{z,\theta}(Z_t|S_t)+\log \nu_{\theta}(S_0).
$$
To finish the proof of this theorem, it suffice to show that $\sigma^2_T=\sup_{\mathcal{M}}CR(\mathcal{M})$. If we can show that
\begin{equation}\label{eq:thm2_1}
 \frac{\partial\eta_{\theta}^{\pi}}{\partial \theta_0}=\mathbb{E}\left[\left\{\sum_{z,a,S_0}c(z|S_0)\cdot p_a(a|z,S_0)\cdot Q^{\pi}(S_0,z,a)+\frac{1}{T}\sum_{j=1}^3 \sum_{t=0}^{T-1}\psi_j^{*}(O_t)\right\}\frac{\partial l(\{O_t\};\theta)}{\partial \theta_0}\right],   
\end{equation}
then according to the property of score function,  $\eta_{\theta}^{\pi}\cdot\mathbb{E}\left[ \frac{\partial l(\{O_t\};\theta)}{\partial \theta_0}\right]=0$, which further indicates that
$$
\frac{\partial\eta_{\theta}^{\pi}}{\partial \theta_0}=\mathbb{E}\left[\left\{\sum_{z,a,S_0}c(z|S_0)\cdot p_a(a|z,S_0)\cdot Q^{\pi}(S_0,z,a)+\frac{1}{T}\sum_{j=1}^3 \sum_{t=0}^{T-1}\psi_j^{*}(O_t)-\eta_{\theta}^{\pi}\right\}\frac{\partial l(\{O_t\};\theta)}{\partial \theta_0}\right].
$$
By Cauchy-Schwarz Inequality, it then follows that
$$
\sigma^2_T\leq \sup_{\mathcal{M}}CR(\mathcal{M})\leq \mathbb{E}\left\{\sum_{z,a,S_0}c(z|S_0)\cdot p_a(a|z,S_0)\cdot Q^{\pi}(S_0,z,a)+\frac{1}{T}\sum_{j=1}^3 \sum_{t=0}^{T-1}\psi_j^{*}(O_t)-\eta_{\theta}^{\pi}\right\}^2=\sigma^2_T,
$$
which implies $\sigma^2_T= \sup_{\mathcal{M}}CR(\mathcal{M})=CR(\mathcal{M}_0)$. (See Lemma 20 of \cite{kallus2022efficiently} for details.)

Therefore, it remains to show that equation \eqref{eq:thm2_1} holds. That is,
$$
\frac{\partial\eta_{\theta}^{\pi}}{\partial \theta_0}=\mathbb{E}\left[\left\{\sum_{z,a,S_0}c(z|S_0)\cdot p_a(a|z,S_0)\cdot Q^{\pi}(S_0,z,a)+\frac{1}{T}\sum_{j=1}^3 \sum_{t=0}^{T-1}\psi_j^{*}(O_t)\right\}\frac{\partial l(\{O_t\};\theta)}{\partial \theta_0}\right].
$$
Notice that in Theorem 1, we’ve proved that
$$
\eta^{\pi}_{\theta_0}=\sum_{s_0}\left[\sum_{t=0}^{+\infty}\gamma^t \sum_{\tau_t,s_{t+1}}r_t\bigg\{\prod_{j=0}^t p_{\theta_0}(s_{j+1},r_j,a_j,z_j|s_j)\bigg\}\cdot\nu_{\theta_0}(s_0)\right],
$$
where $p_{\theta}$ is the probability of a trajectory that follows target policy $\pi$. That is,
$$
p_{\theta_0}(s_{j+1},r_j,a_j,z_j|s_j)= c(z_j|s_j)\cdot p_a(a_j|z_j,s_j)\cdot p_{r,s}(s_{j+1},r_j|a_j,z_j,s_j).
$$
Since $\nabla_{\theta}p_{\theta}=p_{\theta}\nabla_{\theta}\log (p_{\theta})$, we can decompose $\nabla_{\theta}\eta^{\pi}_{\theta_0}$ as
\begin{equation}\label{eq:thm2_2}
 \begin{aligned}
&\nabla_{\theta}\eta^{\pi}_{\theta_0}=\sum_{s_0}\left[\sum_{t=0}^{+\infty}\gamma^t \sum_{\tau_t,s_{t+1}}r_t\bigg\{\prod_{j=0}^t p_{\theta_0}(s_{j+1},r_j,a_j,z_j|s_j)\bigg\}\cdot\bigg\{\sum_{j=0}^t \nabla_{\theta} \log p_{\theta_0}(s_{j+1},r_j,a_j,z_j|s_j)\bigg\}\cdot\nu_{\theta_0}(s_0)\right]\\
&+\sum_{s_0}\left[\sum_{t=0}^{+\infty}\gamma^t \sum_{\tau_t,s_{t+1}}r_t\bigg\{\prod_{j=0}^t p_{\theta_0}(s_{j+1},r_j,a_j,z_j|s_j)\bigg\}\cdot\nu_{\theta_0}(s_0)\cdot \nabla_{\theta}\log \nu_{\theta_0}(s_0)\right]:=L_1+L_2.
\end{aligned}   
\end{equation}
The first line of the equation above can be further derived as 
\[
\begin{aligned}
L_1&= \sum_{s_0}\sum_{j=0}^{+\infty}\Bigg\{\sum_{t=j}^{+\infty}\gamma^t \sum_{\tau_t,s_{t+1}}\bigg[\prod_{k=0}^t p_{\theta_0}(s_{k+1},r_k,a_k,z_k|s_k)\bigg]\Bigg\}\cdot\nabla_{\theta} \log p_{\theta_0}(s_{j+1},r_j,a_j,z_j|s_j)\bigg\}\cdot\nu_{\theta_0}(s_0)\\       
&= \sum_{s_0}\sum_{j=0}^{+\infty}\gamma^j\cdot \sum_{\tau_t,s_{t+1}}\bigg[\prod_{k=0}^j p_{\theta_0}(s_{k+1},r_k,a_k,z_k|s_k)\bigg]\cdot \Bigg\{r_j+\sum_{t=j+1}^{\infty}\gamma^{t-j}\sum_{\tau_{t/j},s_{t+1}} r_t \cdot  \\
&\qquad \qquad\qquad\qquad \bigg[\prod_{k=j+1}^t p_{\theta_0}(s_{k+1},r_k,a_k,z_k|s_k)\bigg]\Bigg\} \cdot \nabla_{\theta} \log p_{\theta_0}(s_{j+1},r_j,a_j,z_j|s_j)\cdot \nu_{\theta_0}(s_0)\\ &=\sum_{s_0}\sum_{j=0}^{+\infty}\gamma^j\sum_{\tau_j,s_{j+1}}\Big\{r_j+\gamma\cdot V^{\pi}(S_{j+1})\Big\}\cdot \bigg[\prod_{k=0}^j p_{\theta_0}(s_{k+1},r_k,a_k,z_k|s_k)\bigg]\cdot \\
&\qquad\qquad\qquad\qquad \nabla_{\theta} \log p_{\theta_0}(s_{j+1},r_j,a_j,z_j|s_j)\cdot \nu_{\theta_0}(s_0).
\end{aligned}
\]
Notice that
$$
\begin{aligned}
&\sum_{s_0}\bigg[\prod_{k=0}^j p_{\theta_0}(s_{k+1},r_k,a_k,z_k|s_k)\bigg]\cdot\nu_{\theta}(s_0)=p_{\theta_0}(s_{j+1},r_j,a_j,z_j|s_j)\cdot p_{t,\theta_0}^{\pi}(s_j)\\         
&= p_{r,s}(s_{j+1},r_j|a_j,z_j,s_j) \cdot p_a(a_j|z_j,s_j)\cdot c(z_j|s_j)\cdot p_{t,\theta_0}^{\pi}(s_j),
\end{aligned}
$$
then by plugging in this expression to $L_1$, we have
$$
\begin{aligned}
L_1&=\sum_{j=0}^{+\infty}\gamma^j\sum_{\tau_j,s_{j+1}}\Big\{r_j+\gamma\cdot V^{\pi}(S_{j+1})\Big\}\cdot p_{\theta_0}(s_{j+1},r_j,a_j,z_j|s_j)\cdot p_{t,\theta_0}^{\pi}(s_j) \cdot \nabla_{\theta} \log p_{\theta_0}(s_{j+1},r_j,a_j,z_j|s_j) \\
&=\sum_{j=0}^{+\infty}\gamma^j\sum_{\tau_j,s_{j+1}}\Big\{r_j+\gamma\cdot V^{\pi}(S_{j+1})\Big\}\cdot p_{\theta_0}(s_{j+1},r_j,a_j,z_j|s_j)\cdot p_{t,\theta_0}^{\pi}(s_j) \cdot \nabla_{\theta} \log p_{s,r,\theta_0}(s_{j+1},r_j|,a_j,z_j,s_j)  \\  &+\sum_{j=0}^{+\infty}\gamma^j\sum_{\tau_j,s_{j+1}}\Big\{r_j+\gamma\cdot V^{\pi}(S_{j+1})\Big\}\cdot p_{\theta_0}(s_{j+1},r_j,a_j,z_j|s_j)\cdot p_{t,\theta_0}^{\pi}(s_j) \cdot \nabla_{\theta} \log p_{a,\theta_0}(a_j|z_j,s_j)  \\ 
&+\sum_{j=0}^{+\infty}\gamma^j\sum_{\tau_j,s_{j+1}}\Big\{r_j+\gamma\cdot V^{\pi}(S_{j+1})\Big\}\cdot p_{\theta_0}(s_{j+1},r_j,a_j,z_j|s_j)\cdot p_{t,\theta_0}^{\pi}(s_j) \cdot \nabla_{\theta} \log c(z_j|s_j)  \\   
&:= (1.1)+(1.2)+(1.3),
\end{aligned}
$$
where the last three lines corresponds to $(1.1)$, $(1.2)$ and $(1.3)$, respectively. We will deal with the detailed expression of these three terms in the next few steps.

Finally, let's focus on the second line of Formula \eqref{eq:thm2_2}. This part, according to our identification result in Theorem \ref{thm:identification}, is equivalent to $\mathbb{E}[V^{\pi}(S_0)\nabla_{\theta}\log \nu_{\theta_0}(S_0)]$. Still, by using the fact that the expectation of a score function is $0$, we have
\begin{equation*}
\begin{aligned}
L_2&=\sum_{s_0}\left[\sum_{t=0}^{+\infty}\gamma^t \sum_{\tau_t,s_{t+1}}r_t\bigg\{\prod_{j=0}^t p_{\theta_0}(s_{j+1},r_j,a_j,z_j|s_j)\bigg\}\cdot\nu_{\theta_0}(s_0)\cdot \nabla_{\theta}\log \nu_{\theta_0}(s_0)\right]\\
&=\mathbb{E}[V^{\pi}(S_0)\nabla_{\theta}\log \nu_{\theta_0}(S_0)]=\mathbb{E}\Big[\big(V^{\pi}(S_0)-\eta^{\pi}_{\theta_0}\big)\nabla_{\theta}\log \nu_{\theta_0}(S_0)\Big]\\
&=\mathbb{E}\Big[\big(V^{\pi}(S_0)-\eta^{\pi}_{\theta_0}\big)S(\bar{O}_{T-1})\Big],
\end{aligned}   
\end{equation*}
where $S(s',r,a,z|s)$ denotes the conditional score function of $(S_{t+1},R_t,A_t,Z_t)|S_t$, and we use $\bar{O}_{T-1}$ to denote all observed data up to stage $T-1$. 

The rest of the proof are organized as follows. In Step 2 to Step 4, we will derive the expression for $(1.1)$, $(1.2)$ and $(1.3)$ respectively. In Step 5, we will focus back on Formula \eqref{eq:thm2_2} to finish the proof of the whole theorem.

\noindent\textbf{Step 2:} Derivation of (1.1).\\
Using the similar trick in deriving $L_2$, we have
$$
\begin{aligned}(1.1)&=\sum_{j=0}^{+\infty}\gamma^j\sum_{\tau_j,s_{j+1}}\Big\{r_j+\gamma\cdot V^{\pi}(s_{j+1})\Big\}\cdot p_{r,s}(s_{j+1},r_j|a_j,z_j,s_j) \cdot p_a(a_j|z_j,s_j)\cdot c(z_j|s_j)\cdot p_{j,\theta_0}^{\pi}(s_j) \\
&\qquad\qquad\qquad\qquad\qquad\qquad\qquad \qquad\qquad\qquad\qquad\qquad\cdot \nabla_{\theta} \log p_{s,r,\theta_0}(s_{j+1},r_j|,a_j,z_j,s_j)\\
&=\sum_{j=0}^{+\infty}\gamma^j\sum_{\tau_j,s_{j+1}}\Big\{r_j+\gamma\cdot V^{\pi}(s_{j+1})-Q^{\pi}(s_j,z_j,a_j)\Big\}\cdot p_{r,s}(s_{j+1},r_j|a_j,z_j,s_j) \cdot p_a(a_j|z_j,s_j)\\
&\qquad\qquad\qquad\qquad\qquad\qquad\qquad \qquad\cdot c(z_j|s_j)\cdot p_{j,\theta_0}^{\pi}(s_j) \cdot \nabla_{\theta} \log p_{s,r,\theta_0}(s_{j+1},r_j|,a_j,z_j,s_j)
\end{aligned}
$$
\[
\begin{aligned}  &=\sum_{j=0}^{+\infty}\gamma^j\sum_{\tau_j,s_{j+1}}\Big\{r_j+\gamma\cdot V^{\pi}(s_{j+1})-Q^{\pi}(s_j,z_j,a_j)\Big\}\cdot p_{r,s}(s_{j+1},r_j|a_j,z_j,s_j) \cdot p_a(a_j|z_j,s_j)\\
&\qquad\qquad\qquad\qquad\qquad\qquad\qquad \qquad\quad\cdot c(z_j|s_j)\cdot p_{j,\theta_0}^{\pi}(s_j) \cdot \nabla_{\theta} \log p_{\theta_0}(s_{j+1},r_j,a_j,z_j|s_j)\\&=\sum_{j=0}^{+\infty}\gamma^j\sum_{\tau_j,s_{j+1}}\Big\{r_j+\gamma\cdot V^{\pi}(s_{j+1})-Q^{\pi}(s_j,z_j,a_j)\Big\}\frac{c(z_j|s_j)}{p_z(z_j|s_j)}\cdot p_{r,s}(s_{j+1},r_j|a_j,z_j,s_j)\\
&\qquad\qquad\qquad\qquad\qquad\cdot p_a(a_j|z_j,s_j)\cdot p_z(z_j|s_j)\cdot p_{j,\theta_0}^{\pi}(s_j) \cdot \nabla_{\theta} \log p_{\theta_0}(s_{j+1},r_j,a_j,z_j|s_j)
\end{aligned}
\]
\[
=\sum_{j=0}^{+\infty}\gamma^j\mathbb{E}\left[\Big\{R_j+\gamma\cdot V^{\pi}(S_{j+1})-Q^{\pi}(S_j,Z_j,A_j)\Big\}\frac{c(Z_j|S_j)}{p_z(Z_j|S_j)}\cdot \frac{p_{j,\theta_0}^{\pi}(S_j)}{p_D(S_j)} \cdot S(S_{j+1},R_j,A_j,Z_j)\right].
\]
Recall that 
$$
\omega^{\pi}(s)=(1-\gamma)\cdot \sum_{t=0}^{+\infty}\gamma^t \frac{p_t^{\pi}(s)}{p_D(s)},\quad \rho(z,s)=\frac{c(z|s)}{p_z(z|s)}.
$$
Then by Markov property, we have
$$
\begin{aligned}
(1.1)&=(1-\gamma)^{-1}\mathbb{E}\left[\Big\{R+\gamma\cdot V^{\pi}(S')-Q^{\pi}(S,Z,A)\Big\}\rho(S,Z)\cdot \omega^{\pi}(S) \cdot S(\bar{O}_{T-1})\right]\\
&:=\mathbb{E}\bigg\{\Big[\frac{1}{T}\sum_t \psi_1(O_t)\Big]S(\bar{O}_{T-1})\bigg\},
\end{aligned}
$$
where we define
\begin{equation}\label{eq:thm2_(1.1)}
\psi_1(O_t)=(1-\gamma)^{-1}\Big\{R_t+\gamma\cdot V^{\pi}(S_{t+1})-Q^{\pi}(S_t,Z_t,A_t)\Big\}\cdot \rho(S_t,Z_t)\cdot \omega^{\pi}(S_t).    
\end{equation}

\noindent\textbf{Step 3:} Derivation of (1.2).
$$
\begin{aligned}
(1.2)&=\sum_{j=0}^{+\infty}\gamma^j\sum_{\tau_j,s_{j+1}}\Big\{r_j+\gamma\cdot V^{\pi}(s_{j+1})\Big\}\cdot p_{\theta_0}(s_{j+1},r_j,a_j,z_j|s_j)\cdot p_{t,\theta_0}^{\pi}(s_j) \cdot \nabla_{\theta} \log p_{a,\theta_0}(a_j|z_j,s_j)\\
&=\sum_{j=0}^{+\infty}\gamma^j\sum_{a_j,z_j,s_j}\mathbb{E}\Big\{\big[R_j+\gamma\cdot V^{\pi}(S_{j+1})\big]|a_j,z_j,s_j\Big\}\cdot p_{a,\theta_0}(a_j|z_j,s_j)\cdot c(z_j|s_j)\cdot p_{t,\theta_0}^{\pi}(s_j)\\
&\qquad\qquad\qquad\qquad\qquad\qquad\qquad\qquad\qquad\qquad\qquad\qquad\qquad\cdot \nabla_{\theta} \log p_{a,\theta_0}(a_j|z_j,s_j)\\
&=\sum_{j=0}^{+\infty}\gamma^j\sum_{a_j,z_j,s_j}Q^{\pi}(s_j,z_j,a_j)\cdot p_{a,\theta_0}(a_j|z_j,s_j)\cdot c(z_j|s_j)\cdot p_{t,\theta_0}^{\pi}(s_j) \cdot \nabla_{\theta} \log p_{a,\theta_0}(a_j|z_j,s_j).
\end{aligned}
$$
Notice that
$$
\begin{aligned}
&\sum_{j=0}^{+\infty}\gamma^j\sum_{a_j,z_j,s_j}\Big\{\sum_{a_j^*}Q^{\pi}(s_j,z_j,a_j^*)\cdot p_{a,\theta_0}(a_j^*|z_j,s_j)\Big\}\cdot p_{a,\theta_0}(a_j|z_j,s_j)\cdot c(z_j|s_j)\cdot p_{t,\theta_0}^{\pi}(s_j)\\
&\qquad\qquad\qquad\qquad\qquad \qquad\qquad\qquad\qquad\qquad\qquad\qquad\qquad\cdot \nabla_{\theta} \log p_{a,\theta_0}(a_j|z_j,s_j)\\
&=\sum_{j=0}^{+\infty}\gamma^j\sum_{z_j,s_j}\Big\{\sum_{a_j^*}Q^{\pi}(s_j,z_j,a_j^*)\cdot p_{a,\theta_0}(a_j^*|z_j,s_j)\Big\}\cdot c(z_j|s_j)\cdot p_{t,\theta_0}^{\pi}(s_j) \\
&\qquad\qquad\qquad\qquad\qquad\cdot \sum_{a_j}p_{a,\theta_0}(a_j|z_j,s_j)\cdot \nabla_{\theta} \log p_{a,\theta_0}(a_j|z_j,s_j)=0,
\end{aligned}
$$
where the last equality holds since the score function has a zero mean.

Combining the formula above with $(1.2)$, we have
$$
\begin{aligned}
(1.2)&=\sum_{j=0}^{+\infty}\gamma^j\sum_{a_j,z_j,s_j}\Big\{Q^{\pi}(s_j,z_j,a_j)-\sum_{a_j^*}Q^{\pi}(s_j,z_j,a_j^*)\cdot p_{a,\theta_0}(a_j^*|z_j,s_j)\Big\}\cdot p_{a,\theta_0}(a_j|z_j,s_j)\cdot c(z_j|s_j)\\
&\qquad\qquad\qquad\qquad\qquad\qquad\qquad\qquad\qquad\qquad\qquad\qquad\qquad\cdot p_{t,\theta_0}^{\pi}(s_j) \cdot \nabla_{\theta} \log p_{a,\theta_0}(a_j|z_j,s_j)\\
&=\sum_{j=0}^{+\infty}\gamma^j\sum_{a_j,z_j,s_j}\Big\{Q^{\pi}(s_j,z_j,a_j)-\sum_{a_j^*}Q^{\pi}(s_j,z_j,a_j^*)\cdot p_{a,\theta_0}(a_j^*|z_j,s_j)\Big\}\\
&\cdot \frac{c(z_j|s_j)}{p_z(z_j|s_j)}\cdot p_{a,\theta_0}(a_j|z_j,s_j)\cdot p_z(z_j|s_j)\cdot p_{t,\theta_0}^{\pi}(s_j) \cdot \nabla_{\theta} \log p_{a,\theta_0}(a_j|z_j,s_j)\\
&=\sum_{j=0}^{+\infty}\gamma^j\sum_{s_{j+1},r_j,a_j,z_j,s_j}\Big\{Q^{\pi}(s_j,z_j,a_j)-\sum_{a_j^*}Q^{\pi}(s_j,z_j,a_j^*)\cdot p_{a,\theta_0}(a_j^*|z_j,s_j)\Big\}\\
&\cdot \rho(z_j,s_j)\cdot p_{s,r,\theta_0}(s_{j+1},r_j|z_j,s_j)\cdot p_{a,\theta_0}(a_j|z_j,s_j)\cdot p_z(z_j|s_j)\cdot p_{t,\theta_0}^{\pi}(s_j) \cdot \nabla_{\theta} \log p_{s,r,a,z,\theta_0}(s_{j+1},r_j,a_j,z_j|s_j)\\
&=\sum_{j=0}^{+\infty}\gamma^j\mathbb{E}\left[\Big\{Q^{\pi}(S,Z,A)-\sum_{a^*}Q^{\pi}(S,Z,a^*)\cdot p_{a,\theta_0}(a^*|Z,S)\Big\}\cdot \rho(S,Z)\cdot \frac{p_{t,\theta_0}^{\pi}(S)}{p_D(S)}\cdot S(S',R,A,Z|S)\right]\\
&=(1-\gamma)^{-1}\cdot \mathbb{E}\left[\omega^{\pi}(S)\rho(S,Z)\Big\{Q^{\pi}(S,Z,A)-\sum_{a^*}Q^{\pi}(S,Z,a^*)\cdot p_{a,\theta_0}(a^*|Z,S)\Big\}\cdot S(S',R,A,Z|S)\right]
\end{aligned}
$$
By defining
\begin{equation}\label{eq:thm2_(1.2)}
\psi_2(O)=(1-\gamma)^{-1}\cdot \omega^{\pi}(S)\cdot \rho(S,Z)\cdot\Big\{Q^{\pi}(S,Z,A)-\sum_{a}Q^{\pi}(S,Z,a)\cdot p_{a}(a|Z,S)\Big\},   
\end{equation}
we can rewrite Formula $(1.2)$ as
\[
(1.2)=\mathbb{E}\left[\bigg\{\frac{1}{T}\sum_t \psi_2(O_t)\bigg\}\cdot S(\bar{O}_{T-1})\right].
\]
\noindent\textbf{Step 4: }Derivation of (1.3).

$$
\begin{aligned}
(1.3)&=\sum_{j=0}^{+\infty}\gamma^j\sum_{\tau_j,s_{j+1}}\Big\{r_j+\gamma\cdot V^{\pi}(S_{j+1})\Big\}\cdot p_{\theta_0}(s_{j+1},r_j,a_j,z_j|s_j)\cdot p_{t,\theta_0}^{\pi}(s_j) \cdot \nabla_{\theta} \log c(z_j|s_j)  \\ &=\sum_{j=0}^{+\infty}\gamma^j\sum_{a_j,z_j,s_j}\mathbb{E}\Big\{\big[R_j+\gamma\cdot V^{\pi}(S_{j+1})\big]|a_j,z_j,s_j\Big\}\cdot p_{a,\theta_0}(a_j|z_j,s_j)\cdot c(z_j|s_j)\cdot p_{t,\theta_0}^{\pi}(s_j) \cdot \nabla_{\theta} \log c(z_j|s_j) \\  &=\sum_{j=0}^{+\infty}\gamma^j\sum_{a_j,z_j,s_j}Q^{\pi}(s_j,z_j,a_j)\cdot p_{a,\theta_0}(a_j|z_j,s_j)\cdot c(z_j|s_j)\cdot p_{t,\theta_0}^{\pi}(s_j) \cdot \nabla_{\theta} \log c(z_j|s_j).
\end{aligned}
$$
Before we proceed, let's calculate $\nabla_{\theta} \log c(z_j|s_j)$ first. According to the definition of $c(z|s)$,
$$
c(z_t|S_t)=\left\{\begin{aligned}
 \frac{p_{1}^A(S_t)-\pi(1|S_t)}{\delta^{A}(S_t)}, &\quad\text{when } z_t=0 \\
\frac{\pi(1|S_t)-p_{0}^A(S_t)}{\delta^{A}(S_t)},  &\quad\text{when } z_t=1
\end{aligned}\right..
$$
Therefore,
$$
\nabla_\theta c(z_t|S_t)=\left\{\begin{aligned}
 \frac{\nabla_{\theta}p_{1}^A(S_t)\delta^{A}(S_t)-(p_{1}^A(S_t)-\pi(1|S_t))\nabla_{\theta}\delta^{A}(S_t)}{[\delta^{A}(S_t)]^2}, &\quad\text{when } z_t=0 \\
\frac{-\nabla_{\theta}p_{0}^A(S_t)\delta^{A}(S_t)-(\pi(1|S_t)-p_{0}^A(S_t))\nabla_{\theta}\delta^{A}(S_t)}{[\delta^{A}(S_t)]^2},  &\quad\text{when } z_t=1
\end{aligned}\right..
$$

According to \cite{wang2018bounded}, 
$$
\nabla_{\theta} p_z^{A_t}(S_t)=\nabla_{\theta} \mathbb{E}\left[A_t|Z_t=z,S_t\right]=\mathbb{E}\left[(A_t-p_z^A(S_t))\cdot S(A_t,Z_t=z|S_t)\right],
$$
and thus
$$
\begin{aligned}
\nabla_{\theta}\delta^{A}(S_t)&=\mathbb{E}\bigg[\frac{2Z_t-1}{p_z(Z_t|S_t)}\left(A_t-\mathbb{E}[A_t|Z_t,S_t]\right)\cdot S(A_t,Z_t|S_t)\Big|S_t\bigg]\\ &=\mathbb{E}\bigg[\left(A_t-p^{A_t}_1(S_t)\right)\cdot S(A_t,Z_t=1|S_t)-\left(A_t-p^{A_t}_0(S_t)\right)\cdot S(A_t,Z_t=0|S_t)\Big|S_t\bigg].
\end{aligned}
$$
Therefore, when $z_t=0$, $\nabla_\theta c(z_t|S_t)$ can be further derived as
$$
\begin{aligned}&\nabla_\theta c(Z_t=0|S_t)=\frac{1}{\delta^{A}(S_t)}\Big\{\nabla_{\theta}p_{1}^A(S_t)-\frac{p_{1}^A(S_t)-\pi(1|S_t)}{\delta^{A}(S_t)}\cdot\nabla_{\theta}\delta^{A}(S_t)\Big\}\\               &=\frac{1}{\delta^{A}(S_t)} \bigg\{\mathbb{E}\Big[(A_t-p_1^{A}(S_t))\cdot S(A_t,Z_t=1|S_t)- c(Z_t=0|S_t)\cdot (A_t-p_1^{A}(S_t))\cdot S(A_t,Z_t=1|S_t)\\
&\qquad\qquad\qquad +c(Z_t=0|S_t)(A_t-p_0^{A}(S_t))\cdot S(A_t,Z_t=0|S_t)\Big]\bigg\}\\ 
&=\frac{1}{\delta^{A}(S_t)} \bigg\{\mathbb{E}\Big[c(Z_t=1|S_t)\cdot (A_t-p_1^{A}(S_t))\cdot S(A_t,Z_t=1|S_t)\\
&\qquad\qquad\qquad+c(Z_t=0|S_t)(A_t-p_0^{A}(S_t))\cdot S(A_t,Z_t=0|S_t)\Big]\bigg\}\\ 
&=\frac{1}{\delta^{A}(S_t)} \bigg\{\mathbb{E}\Big[{\rho(S_t,Z_t)}\cdot (A_t-p_{Z_t}^{A_t}(S_t))\cdot S(A_t,Z_t|S_t)\Big]\bigg\}.\end{aligned}
$$
Likewise, when $z_t=1$,
$$
\begin{aligned}
&\nabla_\theta c(Z_t=1|S_t)=\frac{-1}{\delta^{A}(S_t)}\Big\{\nabla_{\theta}p_{0}^A(S_t)+\frac{\pi(1|S_t)-p_{0}^A(S_t)}{\delta^{A}(S_t)}\cdot\nabla_{\theta}\delta^{A}(S_t)\Big\}\\        &=\frac{-1}{\delta^{A}(S_t)} \bigg\{\mathbb{E}\Big[(A_t-p_0^{A}(S_t))\cdot S(A_t,Z_t=0|S_t)+ c(Z_t=1|S_t)\cdot (A_t-p_1^{A}(S_t))\cdot S(A_t,Z_t=1|S_t)\\
&\qquad\qquad\qquad-c(Z_t=1|S_t)(A_t-p_0^{A}(S_t))\cdot S(A_t,Z_t=0|S_t)\Big]\bigg\}\\ 
&=\frac{-1}{\delta^{A}(S_t)} \bigg\{\mathbb{E}\Big[c(Z_t=1|S_t)\cdot (A_t-p_1^{A}(S_t))\cdot S(A_t,Z_t=1|S_t)\\
&\qquad\qquad\qquad+c(Z_t=0|S_t)(A_t-p_0^{A}(S_t))\cdot S(A_t,Z_t=0|S_t)\Big]\bigg\}\\ 
&=\frac{-1}{\delta^{A}(S_t)} \bigg\{\mathbb{E}\Big[{\rho(S_t,Z_t)}\cdot (A_t-p_{Z_t}^{A_t}(S_t))\cdot S(A_t,Z_t|S_t)\Big]\bigg\}.
\end{aligned}
$$
Combining the results for both $z_t=0$ and $z_t=1$, we can express $\nabla_\theta c(z_t|S_t)$ as
$$
\nabla_\theta c(z_t|S_t)=\frac{(-1)^{z_t}}{\delta^{A}(S_t)}\cdot \mathbb{E}\bigg[{\rho(S_t,Z_t)}\cdot (A_t-p_{Z_t}^{A_t}(S_t))\cdot S(A_t,Z_t|S_t)\bigg].
$$
Therefore,
$$
\begin{aligned}
(1.3)  &=\sum_{j=0}^{+\infty}\gamma^j\sum_{a_j,z_j,s_j}Q^{\pi}(s_j,z_j,a_j)\cdot p_{a,\theta_0}(a_j|z_j,s_j)\cdot c(z_j|s_j)\cdot p_{t,\theta_0}^{\pi}(s_j) \cdot \nabla_{\theta} \log c(z_j|s_j)\\
&=\sum_{j=0}^{+\infty}\gamma^j\sum_{a_j,z_j,s_j}Q^{\pi}(s_j,z_j,a_j)\cdot p_{a,\theta_0}(a_j|z_j,s_j)\cdot p_{t,\theta_0}^{\pi}(s_j) \cdot \nabla_{\theta}c(z_j|s_j)\\ &=\sum_{j=0}^{+\infty}\gamma^j\sum_{a_j,z_j,s_j}Q^{\pi}(s_j,z_j,a_j)\cdot p_{a,\theta_0}(a_j|z_j,s_j)\cdot p_{t,\theta_0}^{\pi}(s_j) \cdot \frac{(-1)^{z_j}}{\delta^{A}(s_j)}\\
&\qquad\qquad\qquad\qquad\cdot \mathbb{E}\bigg[{\rho(S,Z)}\cdot (A-p_{Z}^{A}(S))\cdot S(A,Z|S)\Big|S=s_j\bigg].
\end{aligned}
$$
Notice that 
$$
\sum_{a_j,z_j}Q^{\pi}(s_j,z_j,a_j)\cdot p_{a,\theta_0}(a_j|z_j,s_j) \cdot \frac{(-1)^{z_j}}{\delta^{A}(s_j)}=-\Delta(s_j).
$$
Thus, by replacing this part in the expression of $(1.3)$, we can obtain
$$
\begin{aligned}
(1.3)  &=\sum_{j=0}^{+\infty}\gamma^j\sum_{a_j,z_j,s_j}Q^{\pi}(s_j,z_j,a_j)\cdot p_{a,\theta_0}(a_j|z_j,s_j)\cdot p_{t,\theta_0}^{\pi}(s_j) \cdot \frac{(-1)^{z_j}}{\delta^{A}(s_j)}\\
&\qquad\qquad\qquad\qquad\cdot \mathbb{E}\bigg[{\rho(S,Z)}\cdot (A-p_{Z}^{A}(S))\cdot S(A,Z|S)\Big|S=s_j\bigg]\\  
&=-\sum_{j=0}^{+\infty}\gamma^j\sum_{s_j} p_{t,\theta_0}^{\pi}(s_j) \cdot \Delta(s_j)\cdot\mathbb{E}\bigg[{\rho(S,Z)}\cdot (A-p_{Z}^{A}(S))\cdot S(A,Z|S)\Big|S=s_j\bigg]\\  
&=-(1-\gamma)^{-1}\mathbb{E}\bigg[\omega^{\pi}(S)\cdot{\rho(S,Z)}\cdot (A-p_{Z}^{A}(S))\cdot\Delta(S)\cdot S(A,Z|S)\bigg]\\
&=-(1-\gamma)^{-1}\mathbb{E}\bigg[\omega^{\pi}(S)\cdot{\rho(S,Z)}\cdot (A-p_{Z}^{A}(S))\cdot\Delta(S)\cdot S(S',R,A,Z|S)\bigg]\\
:&=\mathbb{E}\left[\bigg\{\frac{1}{T}\sum_t \psi_3(O_t)\bigg\}\cdot S(\bar{O}_{T-1})\right],
\end{aligned}
$$
where
\begin{equation}\label{eq:thm2_(1.3)}
\psi_3(O_t)=-(1-\gamma)^{-1}\cdot \omega^{\pi}(S_t)\cdot \rho(S_t,Z_t)\cdot(A_t-\mathbb{E}[A_t|S_t,Z_t])\cdot\Delta(S_t).   
\end{equation}

\noindent \textbf{Step 5}: Summary.\\
By change of measure theorem, it can be obtained that
\begin{equation}\label{eq:0toinfty}
    \begin{aligned}  V^{\pi}(S_0)&=\sum_{t=0}^{\infty} \gamma^t\mathbb{E}^{\pi}[R_t|S_0]=\mathbb{E}_{S_t\sim p_t^{\pi}(s)}\bigg[\sum_{t=0}^{\infty} \gamma^t\cdot \mathbb{E}^{\pi}[R_t|S_t]\bigg]\\
    &=\mathbb{E}_{S_t\sim p_{\infty}(s)}\bigg[\sum_{t=0}^{\infty} \gamma^t\cdot \frac{p_t^{\pi}(S_t)}{p_\infty(S_t)}\cdot\mathbb{E}^{\pi}[R_t|S_t]\bigg]=(1-\gamma)^{-1}\mathbb{E}_{S_t\sim p_{\infty}(s)}\bigg[ \omega^{\pi}(S_t)\cdot\mathbb{E}^{\pi}[R_t|S_t]\bigg]\\
    &=(1-\gamma)^{-1}\mathbb{E}_{S_t\sim p_\infty(s)}\left[ \sum_{z_t} \omega^{\pi}(S_t)\cdot c(z_t|S_t)\cdot \mathbb{E}[R_t|Z_t=z_t,S_t]\right].
    \end{aligned}
\end{equation}

Therefore, we can further write 
\begin{equation}\label{eq:thm2_(2)}
\begin{aligned}
L_2&=\mathbb{E}\Big[\big(V^{\pi}(S_0)-\eta^{\pi}_{\theta_0}\big)S(\bar{O}_{T-1})\Big]=\mathbb{E}\bigg[\Big\{(1-\gamma)^{-1}\sum_{z_t} \omega^{\pi}(S_t) c(z_t|S_t) \mathbb{E}[R_t|Z_t=z_t,S_t]-\eta^{\pi}\Big\}\cdot S(\bar{O}_{T-1})\bigg].
\end{aligned}   
\end{equation}

Then, by combining the results in Formula \eqref{eq:thm2_(1.1)}, \eqref{eq:thm2_(1.2)}, \eqref{eq:thm2_(1.3)} and \eqref{eq:thm2_(2)}, we have
\begin{equation*}
 \begin{aligned}
    &\nabla_{\theta}\eta^{\pi}_{\theta_0}=\mathbb{E}\bigg[\Big\{\sum_{j=1}^3 \psi(O_{t})+\sum_{z_t} \omega^{\pi}(S_t) c(z_t|S_t) \mathbb{E}[R_t|Z_t=z_t,S_t]-\eta^{\pi}\Big\}S(\bar{O}_{T-1})\bigg]\\
    &=\mathbb{E}\bigg[\Big\{(1-\gamma)^{-1}\cdot\omega^{\pi}(S_t)\cdot\bigg[\rho(S_t,Z_t)\cdot\Big\{R_t+\gamma\cdot V^{\pi}(S_{t+1})-\mathbb{E}\big[R_t+\gamma\cdot V^{\pi}(S_{t+1})\big|Z_t,S_t\big]\\
    &-\left(A_t-\mathbb{E}\left[A_t\big|Z_t,S_t\right]\right)\cdot \Delta(S_t)\Big\}+\sum_{z_t}c(z_t|S_t)\cdot\mathbb{E}[R_t|z_t,S_t]\bigg]-\eta^{\pi}\Big\}S(\bar{O}_{T-1})\bigg].
\end{aligned}   
\end{equation*}
Therefore, the EIF of $\eta^{\pi}$ is given by
\begin{equation*}
    \begin{aligned}
    \text{EIF}_{\eta^{\pi}}=&(1-\gamma)^{-1}\cdot\omega^{\pi}(S_t)\cdot\bigg[\rho(S_t,Z_t)\cdot\Big\{R_t+\gamma\cdot V^{\pi}(S_{t+1})-\mathbb{E}\big[R_t+\gamma\cdot V^{\pi}(S_{t+1})\big|Z_t,S_t\big]\\
    &-\left(A_t-\mathbb{E}\left[A_t\big|Z_t,S_t\right]\right)\cdot \Delta(S_t)\Big\}+\sum_{z_t}c(z_t|S_t)\cdot\mathbb{E}[R_t|z_t,S_t]\bigg]-\eta^{\pi}.
    \end{aligned}
\end{equation*}
The proof of Theorem 2 is thus complete.

\section{Proof of Theorem \ref{thm:DR}}
We divide the proof of Theorem 3 into two parts. In Section \ref{sec:thm3_1}, we mainly discuss the consistency of $\widehat{\eta}_{\text{DR}}$ under model $\mathcal{M}_1$ or $\mathcal{M}_2$. In Section \ref{sec:thm3_2}, we will prove that $\widehat{\eta}_{\text{DR}}$ is asymptotically normal with variance $\sigma^2_T$, which attains the semiparametric lower bound.
\subsection{Consistency}\label{sec:thm3_1}
\textbf{Scenario 1: }Suppose the models in $\mathcal{M}_1$ and ${p}_a$ are correctly specified.
    
    In this case, $c(z|s)$, $V^{\pi}(S_{t+1})$ and $\Delta(S_t)$ are all correctly specified, since they can all be estimated from $Q^{\pi}(s,z,a)$ and $p_a(a|s,z)$.
    Therefore, 
    $$
    \mathbb{E}\left[\widehat{\eta}^*\right]=\mathbb{E}\left[\widehat{\eta}^{\pi}_{\text{DM}}+(NT)^{-1}\sum_{i,t}\widehat\phi(O_{i,t})\right]=\widehat{\eta}^{\pi}_{\text{DM}}+\mathbb{E}\left[\widehat\phi(O_{t})\right]
    $$
    Now we only need to prove that $\widehat{\eta}^{\pi}_{\text{DM}}$ is consistent to $\eta^\pi$.
    Since $\widehat{c}(z|s)$, $\widehat{p}_a(a|z,s)$ and $\widehat{Q}^{\pi}$ are correctly specified, 
    $$
    \mathbb{E}_{s\sim \widehat{p}_0(s)}\left[\widehat{c}(z|s)\cdot \widehat{p}_a(a|z,s)\cdot \widehat{Q}^{\pi}(s,z,a)\right]-\mathbb{E}_{s\sim \widehat{p}_0(s)}\left[{c}(z|s)\cdot {p}_a(a|z,s)\cdot {Q}^{\pi}(s,z,a)\right]=o_p(1)
    $$
    Also, notice that $\widehat{p}_0(s)$ was set as the estimated initial state distribution. As the sample size goes to infinity, according to WLLN,
    $$
    \mathbb{E}_{s\sim \widehat{p}_0(s)}\left[{c}(z|s)\cdot {p}_a(a|z,s)\cdot {Q}^{\pi}(s,z,a)\right]-\mathbb{E}_{s\sim {p}_0(s)}\left[{c}(z|s)\cdot {p}_a(a|z,s)\cdot {Q}^{\pi}(s,z,a)\right]=o_p(1).
    $$
    Therefore, 
    $$
    \widehat{\eta}^{\pi}_{\text{DM}}\stackrel{P}{\rightarrow}\mathbb{E}_{s\sim {p}_0(s)}\left[{c}(z|s)\cdot {p}_a(a|z,s)\cdot {Q}^{\pi}(s,z,a)\right]=\mathbb{E}_{s\sim {p}_0(s)}\Big[V^{\pi}(S_t)\Big]=\eta^\pi.
    $$
    
    Now it remains to show that $\mathbb{E}\left[\widehat\phi(O_{t})\right]=0$.
    Since $Q^{\pi}(s,z,a), p_a(a|s,z)$ are correctly specified, 
    $$
    \begin{aligned}
    \mathbb{E}\left[\widehat\phi(O_{t})\right]&=(1-\gamma)^{-1}\cdot\mathbb{E}\left[\widehat\omega^{\pi}(S_t)\cdot\widehat\rho(S_t,Z_t)\cdot\Big\{R_t+\gamma\cdot V^{\pi}(S_{t+1})-\mathbb{E}\left[R_t+\gamma\cdot V^{\pi}(S_{t+1})\big|Z_t,S_t\right]\right.\\
    &\qquad\qquad\qquad\qquad\qquad\qquad\qquad\qquad\left.-\left(A_t-\mathbb{E}\left[A_t\big|Z_t,S_t\right]\right)\cdot \Delta(S_t)\Big\}\right]\\   &=(1-\gamma)^{-1}\cdot\mathbb{E}\left[\widehat\omega^{\pi}(S_t)\cdot\widehat\rho(S_t,Z_t)\cdot\mathbb{E}\Big\{R_t+\gamma\cdot V^{\pi}(S_{t+1})-\mathbb{E}\left[R_t+\gamma\cdot V^{\pi}(S_{t+1})\big|Z_t,S_t\right]\right.\\
    &\qquad\qquad\qquad\qquad\qquad\qquad\qquad\qquad\left.-\left(A_t-\mathbb{E}\left[A_t\big|Z_t,S_t\right]\right)\cdot \Delta(S_t)\big|Z_t,S_t\Big\}\Big|S_t,Z_t\right]\\                &=(1-\gamma)^{-1}\cdot\mathbb{E}\left[\widehat\omega^{\pi}(S_t)\cdot\widehat\rho(S_t,Z_t)\cdot\Big\{\mathbb{E}\left[R_t+\gamma\cdot V^{\pi}(S_{t+1})\big|Z_t,S_t\right]\right.\\
    &\qquad\left.-\mathbb{E}\left[R_t+\gamma\cdot V^{\pi}(S_{t+1})\big|Z_t,S_t\right]-\left(\mathbb{E}\left[A_t\big|Z_t,S_t\right]-\mathbb{E}\left[A_t\big|Z_t,S_t\right]\right)\cdot \Delta(S_t)\Big\}\right]\\  &=(1-\gamma)^{-1}\cdot\mathbb{E}\left[\widehat\omega^{\pi}(S_t)\cdot\widehat\rho(S_t,Z_t)\cdot\Big\{ 0-0\cdot \Delta(S_t)\Big\}\right]=0.
    \end{aligned}
    $$
    
\noindent\textbf{Scenario 2: }Suppose the models in $\mathcal{M}_2$ and ${p}_a$ are correctly specified.

In this case, $c(z_t|S_t)$ is also correctly specified since it’s a function of $p_a$ and $\pi$. Then
$$
\begin{aligned}
&\mathbb{E}\left[\widehat{\eta}^{\pi}_{\text{DM}}\right]
=\mathbb{E}_{s\sim p_0(s)}\left[\sum_{z,a}{c}(z|s)\cdot {p}_a(a|s,z)\cdot \widehat{Q}^{\pi}(s,z,a) \right]\\   
&=\mathbb{E}_{S\sim p^\pi_\infty(s),Z\sim p_z}\left[\frac{{c}(Z|S)}{p_z(Z|S)}\cdot\frac{p^{\pi}_0(S)}{p^{\pi}_\infty(S)}\cdot\sum_{a} {p}_a(a|Z,S)\cdot \widehat{Q}^{\pi}(S,Z,a) \right]  =\mathbb{E}_{S\sim p^\pi_\infty(s)}\left[\frac{p^{\pi}_0(S)}{p^{\pi}_\infty(S)}\cdot\widehat{V}^{\pi}(S) \right]. \end{aligned}
$$
The augmentation terms satisfies
$$
\begin{aligned}
\mathbb{E}\left[\widehat\phi(O_{t})\right]&=(1-\gamma)^{-1}\cdot\mathbb{E}\left[\omega^{\pi}(S_t)\cdot\rho(S_t,Z_t)\cdot\Big\{R_t+\gamma\cdot \widehat{V}^{\pi}(S_{t+1})-\sum_{a_t}p_a(a_t|Z_t,S_t)\cdot \widehat{Q}^{\pi}(S_t,Z_t,a_t)\right.\\
&\left.\qquad\qquad\qquad\qquad\qquad\qquad\qquad\qquad\qquad\qquad-\left(A_t-{\mathbb{E}}\left[A_t\big|Z_t,S_t\right]\right)\cdot \widehat{\Delta}(S_t)\Big\}\right]\\           &=(1-\gamma)^{-1}\cdot\mathbb{E}\left[\omega^{\pi}(S_t)\cdot\frac{c(Z_t|S_t)}{p_z(Z_t|S_t)}\cdot\Big\{R_t+\gamma\cdot \widehat{V}^{\pi}(S_{t+1})-\sum_{a_t}p_a(a_t|Z_t,S_t)\cdot \widehat{Q}^{\pi}(S_t,Z_t,a_t)\right.\\
&\left.\qquad\qquad\qquad\qquad\qquad\qquad\qquad\qquad-\left({\mathbb{E}}\left[A_t\big|Z_t,S_t\right]-{\mathbb{E}}\left[A_t\big|Z_t,S_t\right]\right)\cdot \widehat{\Delta}(S_t)\Big\}\right]
\end{aligned}
$$
\[
\begin{aligned}
&=(1-\gamma)^{-1}\cdot\mathbb{E}\left[\omega^{\pi}(S_t)\cdot\frac{c(Z_t|S_t)}{p_z(Z_t|S_t)}\cdot\Big\{R_t+\gamma\cdot \widehat{V}^{\pi}(S_{t+1})-\sum_{a_t}p_a(a_t|Z_t,S_t)\cdot \widehat{Q}^{\pi}(S_t,Z_t,a_t)\Big\}\right]\\       &=(1-\gamma)^{-1}\cdot\mathbb{E}\left[\omega^{\pi}(S_t)\cdot\frac{c(Z_t|S_t)}{p_z(Z_t|S_t)}\cdot R_t\right]+(1-\gamma)^{-1}\cdot\mathbb{E}\left[\omega^{\pi}(S_t)\cdot\frac{c(Z_t|S_t)}{p_z(Z_t|S_t)}\right.\\
&\left.\qquad\qquad\qquad\qquad\qquad\qquad\cdot\Big\{\gamma\cdot \widehat{V}^{\pi}(S_{t+1})-\sum_{a_t}p_a(a_t|Z_t,S_t)\cdot \widehat{Q}^{\pi}(S_t,Z_t,a_t)\Big\}\right].
\end{aligned}
\]

Now it remains to show that 
\begin{equation}\label{eq:intermediate}
\mathbb{E}\left[\widehat{\eta}^{\pi}_{\text{DM}}\right]+(1-\gamma)^{-1}\cdot\mathbb{E}\left[\omega^{\pi}(S_t)\cdot\frac{c(Z_t|S_t)}{p_z(Z_t|S_t)}\cdot\Big\{\gamma\cdot \widehat{V}^{\pi}(S_{t+1})-\sum_{a_t}p_a(a_t|Z_t,S_t)\cdot \widehat{Q}^{\pi}(S_t,Z_t,a_t)\Big\}\right]=0.    
\end{equation}

Since
$$
\begin{aligned}
&(1-\gamma)^{-1}\cdot\mathbb{E}\left[\omega^{\pi}(S_t)\cdot\frac{c(Z_t|S_t)}{p_z(Z_t|S_t)}\cdot\Big\{\gamma\cdot \widehat{V}^{\pi}(S_{t+1})-\sum_{a_t}p_a(a_t|Z_t,S_t)\cdot \widehat{Q}^{\pi}(S_t,Z_t,a_t)\Big\}\right]\\  &=(1-\gamma)^{-1}\cdot\mathbb{E}\left[\omega^{\pi}(S_t)\cdot\frac{{c}(Z_t|S_t)}{p_z(Z_t|S_t)}\cdot \Big\{\gamma\cdot \widehat{V}^{\pi}(S_{t+1})-\sum_{a_t}p_a(a_t|Z_t,S_t)\cdot \widehat{Q}^{\pi}(S_t,Z_t,a_t)\Big\}\right]\\    
&=(1-\gamma)^{-1}\cdot\mathbb{E}_{S\sim p_\infty}\left[\omega^{\pi}(S)\cdot\frac{{c}(Z|S)}{p_z(Z|S)}\cdot\gamma\cdot \widehat{V}^{\pi}(S^*)\right]\\
&\qquad-(1-\gamma)^{-1}\cdot\mathbb{E}_{S\sim p_\infty}\left[\omega^{\pi}(S)\cdot\frac{{c}(Z|S)}{p_z(Z|S)}\cdot\sum_{a}p_a(a|Z,S)\cdot \widehat{Q}^{\pi}(S,Z,a)\right]\\       
&=(1-\gamma)^{-1}\cdot\mathbb{E}_{S\sim p_\infty}\left[\omega^{\pi}(S)\cdot\frac{{c}(Z|S)}{p_z(Z|S)}\cdot\gamma\cdot \widehat{V}^{\pi}(S^*)\right]\\
&\qquad-(1-\gamma)^{-1}\cdot\mathbb{E}_{S\sim p_\infty}\left[\omega^{\pi}(S)\sum_{a,z}{c}(z|S)\cdot p_a(a|z,S)\cdot \widehat{Q}^{\pi}(S,z,a)\right]\\                 
&=(1-\gamma)^{-1}\cdot\mathbb{E}_{S\sim p_\infty}\left[\omega^{\pi}(S)\cdot\frac{{c}(Z|S)}{p_z(Z|S)}\cdot\gamma\cdot \widehat{V}^{\pi}(S^*)\right]-(1-\gamma)^{-1}\cdot\mathbb{E}_{S\sim p_\infty}\left[\omega^{\pi}(S) \cdot \widehat{V}^{\pi}(S)\right].
\end{aligned}
$$
By plugging in the definition for $\omega^{\pi}(s)$, we have 

$$
\begin{aligned}
&(1-\gamma)^{-1}\cdot\mathbb{E}_{S\sim p_\infty}\left[\omega^{\pi}(S)\cdot\frac{{c}(Z|S)}{p_z(Z|S)}\cdot\gamma\cdot \widehat{V}^{\pi}(S^*)\right]-(1-\gamma)^{-1}\cdot\mathbb{E}_{S\sim p_\infty}\left[\omega^{\pi}(S) \cdot \widehat{V}^{\pi}(S)\right]\\  &=\mathbb{E}_{S\sim p_\infty}\left[\sum_{t=0}^\infty \gamma^t\frac{p_t^{\pi}(S)}{p_{\infty}(S)}\cdot\frac{{c}(Z|S)}{p_z(Z|S)}\cdot\gamma\cdot \widehat{V}^{\pi}(S^*)\right]-(1-\gamma)^{-1}\cdot\mathbb{E}_{S^*\sim p_\infty}\left[\omega^{\pi}(S^*) \cdot \widehat{V}^{\pi}(S^*)\right]\\ &=\mathbb{E}_{S\sim p_\infty}\left[\sum_{t=0}^\infty \gamma^{t+1}\cdot\frac{p_t^{\pi}(S)}{p_{\infty}(S)}\cdot\frac{{c}(Z|S)}{p_z(Z|S)}\cdot\widehat{V}^{\pi}(S^*)\right]-\mathbb{E}_{S^*\sim p_\infty}\left[\sum_{t=0}^\infty \gamma^{t}\cdot\frac{p_t^{\pi}(S^*)}{p_{\infty}(S^*)}\cdot \widehat{V}^{\pi}(S^*)\right]\\ &=\mathbb{E}_{S\sim p_\infty}\left[\sum_{t=0}^\infty \gamma^{t+1}\cdot\frac{p_t^{\pi}(S)}{p_{\infty}(S)}\cdot\frac{{c}(Z|S)}{p_z(Z|S)}\cdot\widehat{V}^{\pi}(S^*)\right]-\mathbb{E}_{S^*\sim p_\infty}\left[\sum_{t=1}^\infty \gamma^{t}\cdot\frac{p_t^{\pi}(S^*)}{p_{\infty}(S^*)}\cdot \widehat{V}^{\pi}(S^*)\right]\\
&\qquad\qquad\qquad\qquad\qquad\qquad\qquad\qquad\qquad\qquad\qquad\qquad-\mathbb{E}_{S^*\sim p_\infty}\left[\frac{p_0^{\pi}(S^*)}{p_{\infty}(S^*)}\cdot \widehat{V}^{\pi}(S^*)\right]\\ 
&=\mathbb{E}_{S\sim p_\infty}\left[\sum_{t=0}^\infty \gamma^{t+1}\cdot\bigg\{\frac{p_t^{\pi}(S)}{p_{\infty}(S)}\cdot\frac{{c}(Z|S)}{p_z(Z|S)}-\frac{p_{t+1}^{\pi}(S^*)}{p_{\infty}(S^*)}\bigg\}\cdot \widehat{V}^{\pi}(S^*)\right]-\mathbb{E}_{S^*\sim p_\infty}\left[\frac{p_0^{\pi}(S^*)}{p_{\infty}(S^*)}\cdot \widehat{V}^{\pi}(S^*)\right].
\end{aligned}
$$
Since $\mathbb{E}\left[\widehat{\eta}^{\pi}_{\text{DM}}\right]=\mathbb{E}_{S\sim p^\pi_\infty(s),Z\sim p_z}\left[\frac{p^{\pi}_0(S)}{p^{\pi}_\infty(S)}\cdot\widehat{V}^{\pi}(S) \right]$can be cancelled with the second term of the last expression, all we need to prove is that
$$
\mathbb{E}_{S\sim p_\infty}\left[\bigg\{\frac{p_t^{\pi}(S)}{p_{\infty}(S)}\cdot\frac{{c}(Z|S)}{p_z(Z|S)}-\frac{p_{t+1}^{\pi}(S^*)}{p_{\infty}(S^*)}\bigg\}\cdot \widehat{V}^{\pi}(S^*)\right]=0, \quad \text{for any t}\in \mathbb{N}.
$$
This statement holds naturally because, for any $t$,
$$
\begin{aligned}
&\mathbb{E}_{S\sim p_\infty}\left[\frac{p_t^{\pi}(S)}{p_{\infty}(S)}\cdot\frac{{c}(Z|S)}{p_z(Z|S)}\cdot \widehat{V}^{\pi}(S^*)\right]=\mathbb{E}_{S\sim p_t^{\pi}(S)}\left[\frac{{c}(Z|S)}{p_z(Z|S)}\cdot \widehat{V}^{\pi}(S^*)\right]\\&=\mathbb{E}_{S\sim p_t^{\pi}(S)}\left[\sum_z{c}(z|S)\cdot \widehat{V}^{\pi}(S^*)\right]=\mathbb{E}_{S\sim p_t^{\pi}(S)}\left[\left({c}(z=1|S)+{c}(z=0|S)\right)\cdot \widehat{V}^{\pi}(S^*)\right]\\&=\mathbb{E}_{S\sim p_t^{\pi}(S)}\left[\widehat{V}^{\pi}(S^*)\right]=\mathbb{E}_{S^*\sim p_{t+1}^{\pi}(S^*)}\left[\widehat{V}^{\pi}(S^*)\right]=\mathbb{E}_{S\sim p_\infty}\left[\frac{p_{t+1}^{\pi}(S^*)}{p_{\infty}(S^*)}\cdot \widehat{V}^{\pi}(S^*)\right].
\end{aligned}^*
$$
Therefore, under $\mathcal{M}_2$, i.e. when $p_a, p_z,\omega^{\pi}$ are correctly specified, $\widehat{\eta}_{\text{DR}}$ is consistent to $\eta^\pi$.

Based on the results in both scenarios, the double robustness of our estimator $\widehat{\eta}_{\text{DR}}$ thus holds.

\subsection{Asymptotic Normality}\label{sec:thm3_2}

For the simplicity of notations, we drop the subscript and denote our final DR estimator as $\widehat{\eta}$. First, let's define the oracle estimator $\widehat{\eta}^*$ as 
\[
\widehat{\eta}^*:={\eta}_{\text{DM}}^*+(NT)^{-1}\sum_{i,t}{\phi}^*(O_{i,t}), 
\]
where we use the star sign in superscript to indicate that all of the models in ${\eta}_{\text{DM}}$ and $\phi$ are substituted by their ground truth. To be more clear, we define ${\eta}_{\text{DM}}^*:={\eta}_{\text{DM}}^{\pi}({p}_a,{Q}^{\pi})$, and $\phi^*(O_{i,t}):=\phi(O_{i,t},{p}_a,{Q}^{\pi},{p}_z,{\omega}^\pi)$. Similarly, to emphasize the dependence of $\widehat\eta_{\text{DM}}^{\pi}$ and $\widehat\phi$ on nuisance functions, we denote $\widehat{\eta}_{\text{DM}}^{\pi}:={\eta}_{\text{DM}}^{\pi}(\widehat{p}_a,\widehat{Q}^{\pi})$, and $\widehat\phi(O_{i,t}):=\phi(O_{i,t},\widehat{p}_a,\widehat{Q}^{\pi},\widehat{p}_z,\widehat{\omega}^\pi)$.

The proof of Section \ref{sec:thm3_2} is decomposed to two steps. In Step 1, we will show that our DR estimator, $\widehat{\eta}$, is asymptotically equivalent to the oracle estimator $\widehat{\eta}^*$ by proving $\|\widehat{\eta}-\widehat{\eta}^*\|=o_p(N^{-\frac{1}{2}})$. In Step 2, we will illustrate the asymptotic normality of our oracle estimator $\widehat{\eta}^*$, and prove that the asymptotic variance indeed reaches the semiparametric efficiency bound.

\noindent\textbf{Step 1:}  show that $\|\widehat{\eta}-\widehat{\eta}^*\|=o_p(N^{-\frac{1}{2}})$.

According to the assumptions in Theorem 2, all of the nuisance functions, i.e. $\widehat{p}_z$, $\widehat{p}_a$, $\widehat{Q}^{\pi}$ and $\widehat{\omega}^{\pi}$, converge in $L_2$-norm to their oracle values with rates no smaller than $1/4$. That is, there exist a constant $\alpha\geq1/4$, such that
\[
\begin{aligned}
& \sqrt{\mathbb{E}_{s\sim p_{\infty}}\mathbb{E}\big|\widehat{p}_z(z|s)-{p}_z(z|s)\big|^2}=O_p(N^{-\alpha}),\quad \sqrt{\mathbb{E}_{s\sim p_{\infty}}\mathbb{E}\big|\widehat{p}_a(a|z,s)-{p}_a(a|z,s)\big|^2}=O_p(N^{-\alpha}),\\
&\sqrt{\mathbb{E}_{s\sim p_{\infty}}\mathbb{E}\big|\widehat{Q}^{\pi}(s,z,a)-{Q}^{\pi}(s,z,a)\big|^2}=O_p(N^{-\alpha}),\quad \sqrt{\mathbb{E}_{s\sim p_{\infty}}\mathbb{E}\big|\widehat{\omega}^{\pi}(s)-{\omega}^{\pi}(s)\big|^2}=O_p(N^{-\alpha}).
\end{aligned}
\]
Based on definition of oracle estimator, we can decompose the difference between $\widehat{\eta}$ and $\widehat{\eta}^*$ as
\begin{equation*}
\begin{aligned}
\widehat{\eta}-\widehat{\eta}^* &=\Big[{\eta}^{\pi}_{\text{DM}}(\widehat{p}_a,\widehat{Q}^{\pi})-{\eta}^{\pi}_{\text{DM}}({p}_a,{Q}^{\pi})\Big] +\frac{1}{NT}\sum_{i,t}\Big[\phi(O_{i,t},\widehat{p}_a,\widehat{Q}^{\pi},\widehat{p}_z,\widehat{\omega}^\pi)-{\phi}(O_{i,t},{p}_a,{Q}^{\pi},{p}_z,{\omega}^\pi)\Big]\\
:&=\Delta^{(1)}+\Delta^{(2)}+\Delta^{(3)},
\end{aligned}    
\end{equation*}
where we define
\begin{equation}\label{eq:Thm2_1}
    \begin{aligned}
    \Delta^{(1)}=&\Big[{\eta}^{\pi}_{\text{DM}}({p}_a,\widehat{Q}^{\pi})-{\eta}^{\pi}_{\text{DM}}({p}_a,{Q}^{\pi})\Big] +\frac{1}{NT}\sum_{i,t}\Big[\phi(O_{i,t},{p}_a,\widehat{Q}^{\pi},{p}_z,{\omega}^\pi)-{\phi}(O_{i,t},{p}_a,{Q}^{\pi},{p}_z,{\omega}^\pi)\Big],\\
\Delta^{(2)}
=&\Big[{\eta}^{\pi}_{\text{DM}}(\widehat{p}_a,\widehat{Q}^{\pi})-{\eta}^{\pi}_{\text{DM}}({p}_a,\widehat{Q}^{\pi})\Big] +\frac{1}{NT}\sum_{i,t}\Big[\phi(O_{i,t},\widehat{p}_a,\widehat{Q}^{\pi},{p}_z,{\omega}^\pi)-\phi(O_{i,t},{p}_a,\widehat{Q}^{\pi},{p}_z,{\omega}^\pi)\Big],\\
\Delta^{(3)}=&\frac{1}{NT}\sum_{i,t}\Big[\phi(O_{i,t},\widehat{p}_a,\widehat{Q}^{\pi},\widehat{p}_z,\widehat{\omega}^\pi)-{\phi}(O_{i,t},\widehat{p}_a,\widehat{Q}^{\pi},{p}_z,{\omega}^\pi)\Big].
    \end{aligned}
\end{equation}
To finish the proof of this step, it suffice to show that the three terms in \eqref{eq:Thm2_1} are all $o_p(N^{-1/2})$ for any $\widehat{p}_z$, $\widehat{p}_a$, $\widehat{Q}^{\pi}$ and $\widehat{\omega}^{\pi}$ in a neighborhood of their true values.

For the brevity of content, we will only give a sketch of the proof and detail on just one term for illustration purpose. 

Let's first consider $\Delta^{(1)}$. According to the double robustness property that will be proved in Theorem \ref{thm:DR}, for any $\widehat{Q}^{\pi}$ that may be misspecified in $\Delta^{(1)}$, we have $\mathbb{E}[\Delta^{(1)}]=0$. Define $\mathcal{Q}^{\pi}$ as a neighborhood of $Q^{\pi}$. We construct an empirical operator $\mathbb{G}_N$ as
\[
\mathbb{G}_N(\widehat{Q}^{\pi})=\sqrt{N}\Big\{\Delta^{(1)}(p_a,\widehat{Q}^{\pi},{p}_z,{\omega}^\pi)-\mathbb{E}\big[\Delta^{(1)}(p_a,\widehat{Q}^{\pi},{p}_z,{\omega}^\pi)\big]\Big\},
\]
Define a function class $\mathcal{F}_{Q}=\big\{\eta_{\text{DR}}(p_a,\widehat{Q}^{\pi},{p}_z,{\omega}^\pi)-\eta_{\text{DR}}(p_a,{Q}^{\pi},{p}_z,{\omega}^\pi): \widehat{Q}^{\pi}\in\mathcal{Q}^{\pi}\big\}$ with $\eta_{\text{DR}}(p_a,{Q}^{\pi},{p}_z,{\omega}^\pi)= \sum_{z_0,a_0}{c}(z_{0}|S_{0}){p}_a(a_{0}|z_{0},S_{0}) {Q}^{\pi}(S_{0},z_{0},a_{0})+ \frac{1}{T}\sum_{t=1}^T \phi(O_t,p_a,{Q}^{\pi},{p}_z,{\omega}^\pi)$. It can be proved that $\mathcal{F}_{Q}$ is a VC-type class. By applying the Maximal Inequality specialized to VC type classes \citep{chernozhukov2014gaussian}, one can show that $\sup_{\widehat{Q}^{\pi}\in\mathcal{Q}^{\pi}}\|\mathbb{G}_N(\widehat{Q}^{\pi})\|=o_p(1)$, which further indicates that
\[
\sup_{\widehat{Q}^{\pi}\in\mathcal{Q}^{\pi}}\left\|\Delta^{(1)}(p_a,\widehat{Q}^{\pi},{p}_z,{\omega}^\pi)\right\|=\sup_{\widehat{Q}^{\pi}\in\mathcal{Q}^{\pi}}\left\|\Delta^{(1)}(p_a,\widehat{Q}^{\pi},{p}_z,{\omega}^\pi)-\mathbb{E}\big[\Delta^{(1)}(p_a,\widehat{Q}^{\pi},{p}_z,{\omega}^\pi)\big]\right\|=o_p(N^{-1/2}).
\]

Therefore, as long as $\widehat{Q}^{\pi}$ converges in $L_2$ norm to its oracle value with rate $\alpha>1/4$, we have $\Delta^{(1)}=o_p(N^{-1/2})$. [See \cite{van1996weak} for details.]

Next, let's consider $\Delta^{(2)}$. This term can be further decomposed as
\[
\small
\begin{aligned}
&\Delta^{(2)}=\Big[{\eta}^{\pi}_{\text{DM}}(\widehat{p}_a,\widehat{Q}^{\pi})-{\eta}^{\pi}_{\text{DM}}({p}_a,\widehat{Q}^{\pi})-{\eta}^{\pi}_{\text{DM}}(\widehat{p}_a,{Q}^{\pi})+\widehat{\eta}^{\pi}_{\text{DM}}({p}_a,{Q}^{\pi})\Big]\\
&+\frac{1}{NT}\sum_{i,t}\Big[\phi(O_{i,t},\widehat{p}_a,\widehat{Q}^{\pi},{p}_z,{\omega}^\pi)-\phi(O_{i,t},{p}_a,\widehat{Q}^{\pi},{p}_z,{\omega}^\pi)-\phi(O_{i,t},\widehat{p}_a,{Q}^{\pi},{p}_z,{\omega}^\pi)+\phi(O_{i,t},{p}_a,{Q}^{\pi},{p}_z,{\omega}^\pi)\Big]\\
&+\Big[{\eta}^{\pi}_{\text{DM}}(\widehat{p}_a,{Q}^{\pi})-\widehat{\eta}^{\pi}_{\text{DM}}({p}_a,{Q}^{\pi})\Big] +\frac{1}{NT}\sum_{i,t}\Big[\phi(O_{i,t},\widehat{p}_a,{Q}^{\pi},{p}_z,{\omega}^\pi)-\widehat\phi(O_{i,t},{p}_a,{Q}^{\pi},{p}_z,{\omega}^\pi)\Big].
\end{aligned}
\]
The last line of the equation above, akin to $\Delta^{(1)}$, can be proved to be $o_p(N^{-1/2})$ under the $L_2$ convergence assumption of $\widehat{Q}^{\pi}$. The first two lines, after some manipulations, can be bounded by the product of two nuisance function error terms. To be more specific, let's consider the first line and prove that it is bound with order $o_p(N^{-1/2})$.
\[
\begin{aligned}
& \Big|{\eta}^{\pi}_{\text{DM}}(\widehat{p}_a,\widehat{Q}^{\pi})-{\eta}^{\pi}_{\text{DM}}({p}_a,\widehat{Q}^{\pi})-{\eta}^{\pi}_{\text{DM}}(\widehat{p}_a,{Q}^{\pi})+\widehat{\eta}^{\pi}_{\text{DM}}({p}_a,{Q}^{\pi})\Big|\\
&=\bigg|\frac{1}{N}\sum_{i=1}^N\sum_{z,a}\Big\{\widehat{c}(z|S_{i,0}) \widehat{p}_a(a|z,S_{i,0}) \widehat{Q}^{\pi}(S_{i,0},z,a)-{c}(z|S_{i,0}) {p}_a(a|z,S_{i,0}) \widehat{Q}^{\pi}(S_{i,0},z,a)\Big\}\\
&\quad -\frac{1}{N}\sum_{i=1}^N\sum_{z,a}\Big\{\widehat{c}(z|S_{i,0}) \widehat{p}_a(a|z,S_{i,0}) {Q}^{\pi}(S_{i,0},z,a)-{c}(z|S_{i,0}) {p}_a(a|z,S_{i,0}) {Q}^{\pi}(S_{i,0},z,a) \Big\}\bigg|\\
&=\bigg|\frac{1}{N}\sum_{i=1}^N\sum_{z,a}\Big\{\Big[\widehat{c}(z|S_{i,0}) \widehat{p}_a(a|z,S_{i,0})-{c}(z|S_{i,0}){p}_a(a|z,S_{i,0})\Big]\cdot \Big[ \widehat{Q}^{\pi}(S_{i,0},z,a)- {Q}^{\pi}(S_{i,0},z,a)\Big]\Big\}\bigg|\\
&\leq \frac{1}{N}\sum_{i=1}^N\sum_{z,a}\Big|\widehat{c}(z|S_{i,0}) \widehat{p}_a(a|z,S_{i,0})-{c}(z|S_{i,0}){p}_a(a|z,S_{i,0})\Big|\cdot \Big| \widehat{Q}^{\pi}(S_{i,0},z,a)- {Q}^{\pi}(S_{i,0},z,a)\Big|\\
&=o_p(N^{-1/4})\cdot o_p(N^{-1/4})= o_p(N^{-1/2}),
\end{aligned}
\]
where the second last equality is derived from the $L_2$ convergence of $\widehat{p}_a$ and $\widehat{Q}^{\pi}$ with rate $\alpha\geq 1/4$.

The proof of $\Delta^{(3)}$ is also similar to other terms, which is thus omitted. A similar proof can be found in \cite{shi2022off}. Therefore, $\|\widehat{\eta}-\widehat{\eta}^*\|=o_p(N^{-\frac{1}{2}})$.

\noindent\textbf{Step 2:}  show that $\sqrt{N}(\widehat{\eta}^*-\eta^{\pi})\stackrel{d}{\rightarrow} \mathcal{N}(0,\sigma^2_T)$, where $\sigma^2_T$ is the efficiency bound.\\
Since $\widehat{\eta}^*$ is the average of $N$ i.i.d. data trajectories, by standard CLT, the oracle estimator $\widehat{\eta}^*$ is asymptotically normal with variance 
\begin{eqnarray*}
{\sigma}^2_T&=&\text{Var}\bigg\{\sum_{z,a}{c}(z|S_0)\cdot {p}_a(a|z,S_0)\cdot {Q}^{\pi}(S_0,z,a)+\frac{1}{T}\sum_{t=1}^T\phi(O_t)\bigg\}\\
&=&\text{Var}\bigg\{ V^{\pi}(S_0)+\frac{1}{T}\sum_{t=1}^T\phi(O_t) \bigg\}=\text{Var}\bigg\{ V^{\pi}(S_0)\bigg\}+\frac{1}{T^2}\sum_{t=1}^T\text{Var}\bigg\{\phi(O_t)\bigg\},
\end{eqnarray*}
where the second equality is due to \eqref{eq:relation_Q_V} and the last equality is due to the Markov and conditional mean independence assumptions. 
It follows directly from the proof of Theorem 2 that $\sigma^2_T=CR(\mathcal{M})$. 

Therefore, by combining the claims in Step 1 and Step 2, we have
\[
\sqrt{N}(\widehat{\eta}_{\text{DR}}-\eta^{\pi})\stackrel{d}{\rightarrow} \mathcal{N}(0,\sigma^2_T).
\]
The proof of Theorem 3 is thus complete.

\section{Proof of Theorem \ref{thm:POMDP_identification}}
In this section, we detail the identification proof under confounded POMDP. For any $g_Q\in \mathcal{G}_Q$, 
\[
\begin{aligned}
   0&=\mathbb{E}\Big\{R+\gamma \sum_{z,a}c(z|O')\cdot p_a(a|z,O')\cdot g_Q(F',z,a)-g_Q(F,z,a)\Big| H,Z,A\Big\}\\
   &= \mathbb{E}\bigg\{\mathbb{E}\big[R+\gamma \sum_{z,a}c(z|O')\cdot p_a(a|z,O')\cdot g_Q(F',z,a)-g_Q(F,z,a)\big| S,H,Z,A\big]\Big| H,Z,A\bigg\},
\end{aligned}
\]
where the first equality holds by the definition of learnable future-dependent Q function, and the second equality holds by the law of total expectation.

Since $H\!\perp\!\!\!\perp F'|(S,Z,A)$, we have
\[
\begin{aligned}
   0= \mathbb{E}\bigg\{\mathbb{E}\big[R+\gamma \sum_{z,a}c(z|O')\cdot p_a(a|z,O')\cdot g_Q(F',z,a)-g_Q(F,z,a)\big| S,Z,A\big]\Big| H,Z,A\bigg\}.
\end{aligned}
\]
Then by the invertibility assumption, it holds almost surely that
\[
\mathbb{E}\Big[R+\gamma \sum_{z,a}c(z|O')\cdot p_a(a|z,O')\cdot g_Q(F',z,a)-g_Q(F,z,a)\big| S,Z,A\Big]=0.
\]
According to the derivation details of importance sampling estimator in Formula (\ref{eq:0toinfty}), the aggregated value $\eta^{\pi}$ can be written as
\[
\eta^{\pi}=(1-\gamma)^{-1}\mathbb{E}_{S\sim p_{\infty}} \Big[\omega^{\pi}(S)\cdot\mathbb{E}\big\{\rho(Z,O)\cdot R|S\big\}\Big].
\]
In Formula (\ref{eq:intermediate}), since $\widehat{Q}$ is misspecified, one can substitute $\widehat{Q}$ to $g_Q$ and the equality still holds. That is,
\begin{equation}\label{eq:itermediate2}
\begin{aligned}
0&=\mathbb{E}_{f\sim\nu_F}\bigg[\sum_{z,a}c(z|o)p_a(a|z,o)g_Q(f,z,a)\bigg]+(1-\gamma)^{-1}\mathbb{E}\bigg[\omega^{\pi}(S)\cdot\rho(Z,O)\cdot\\
&\qquad\Big\{\gamma\cdot \sum_{z,a}c(z|O')p_a(a|z,O')g_Q(F',z,a)-\sum_{a}p_a(a|Z,O)\cdot g_Q(F,Z,a)\Big\}\bigg].
\end{aligned}   
\end{equation}
Therefore, by plugging in the expression in \eqref{eq:itermediate2}, we have
\[
\begin{aligned}
&\eta^{\pi}-\mathbb{E}_{f\sim \nu_F}\Big[\sum_{z,a} c(z|o)p_a(a|z,o)g_Q(f,z,a)\Big]\\
=&(1-\gamma)^{-1}\mathbb{E}_{s\sim p_{\infty}} \Big[\omega^{\pi}(s)\cdot\mathbb{E}\big\{\rho(Z,O)\cdot R|S=s\big\}\Big]-\mathbb{E}_{f\sim \nu_F}\Big[\sum_{z,a} c(z|o)p_a(a|z,o)g_Q(f,z,a)\Big]\\\
=&(1-\gamma)^{-1}\mathbb{E}\bigg[\omega^{\pi}(S)\rho(Z,O)\Big\{R+\gamma\sum_{z,a}c(z|O')p_a(a|z,O')g_Q(F',z,a)-\sum_{a}p_a(a|Z,O)\cdot g_Q(F,Z,a)\Big\}\bigg]\\
=& (1-\gamma)^{-1}\mathbb{E}\bigg[\omega^{\pi}(S)\rho(Z,O)\Big\{R+\gamma\sum_{z,a}c(z|O')p_a(a|z,O')g_Q(F',z,a)- g_Q(F,Z,A)\Big\}\bigg]=0,
\end{aligned}
\]
where the last equality holds by the overlap condition and the definition of learnable future-dependent Q function. Therefore, $\eta^\pi$ can be written as a function of observed data, which is indeed identifiable. The identification expression for $\eta^{\pi}$ is given by
\[
\eta^{\pi}=\mathbb{E}_{f\sim \nu_F}\Big[\sum_{z,a} c(z|o)p_a(a|z,o)g_Q(f,z,a)\Big],
\]
which finishes the proof of this theorem.

\end{document}